%% file: arxiv_v2.tex
\documentclass[acmsmall, screen]{acmart}

\AtBeginDocument{%
  }

\renewcommand\footnotetextcopyrightpermission[1]{} 

\acmPrice{15.00}
\acmISBN{978-1-4503-XXXX-X/18/06}

\usepackage{wrapfig}
\usepackage{caption}
\usepackage{subcaption}
\usepackage{enumitem}
\usepackage{flushend}
\usepackage{balance}
\usepackage{lineno}
\usepackage{algorithm}
\usepackage{algorithmic}
\usepackage{graphicx}
\usepackage{textcomp}
\usepackage{xcolor}
\usepackage{colortbl}
\usepackage{amsthm}
\usepackage{url}
\usepackage{float}
\usepackage{multirow}
\usepackage{multicol}
\usepackage{color}
\usepackage{bm}
\usepackage{bbm}

\usepackage{tikz}
\usepackage{forest}

\newtheorem{definition}{Definition}

\usepackage{pifont}

\def \st{\;\text{s.t.}\;}

\newcommand{\bh}{\mathbf{h}}

\usepackage{xcolor,pifont}

\newcommand*\colourcheck[1]{%
  \expandafter\newcommand\csname #1check\endcsname{\textcolor{#1}{\ding{52}}}%
}
\colourcheck{blue}
\colourcheck{green}
\colourcheck{red}

\newcommand*\colourwrong[1]{%
  \expandafter\newcommand\csname #1wrong\endcsname{\textcolor{#1}{\ding{55}}}%
}
\colourwrong{blue}
\colourwrong{green}
\colourwrong{red}

\usepackage{array}
\newcolumntype{L}[1]{>{\raggedright\let\newline\\\arraybackslash\hspace{0pt}}m{#1}}
\newcolumntype{C}[1]{>{\centering\let\newline  \\\arraybackslash\hspace{0pt}}m{#1}}
\newcolumntype{R}[1]{>{\raggedleft\let\newline \\\arraybackslash\hspace{0pt}}m{#1}}
\DeclareMathOperator*{\argmin}{argmin}

\usepackage[utf8]{inputenc} % allow utf-8 input
\usepackage[T1]{fontenc}    % use 8-bit T1 fonts
\usepackage{hyperref}       % hyperlinks
\usepackage{url}            % simple URL typesetting
\usepackage{booktabs}       % professional-quality tables
\usepackage{amsfonts}       % blackboard math symbols
\usepackage{nicefrac}       % compact symbols for 1/2, etc.
\usepackage{microtype}      % microtypography
\usepackage{xcolor}         % colors

\begin{document}

%%
%% The "title" command has an optional parameter,
%% allowing the author to define a "short title" to be used in page headers.
\title{Knowledge Editing for Large Language Models: A Survey}

% \author{\small Song Wang, Yaochen Zhu, Haochen Liu, Zaiyi Zheng, Chen Chen, Jundong Li} 
% \affiliation{%
%   \institution{\\University of Virginia}
%   \institution{\\\{uqp4qh, jundong\}@virginia.edu}
%   \country{}
% }

\author{Song Wang}
\affiliation{%
\institution{University of Virginia}
\city{Charlottesville}
\state{Virginia}
\country{USA}
}
\email{sw3wv@virginia.edu}

\author{Yaochen Zhu}
\affiliation{%
\institution{University of Virginia}
\city{Charlottesville}
\state{Virginia}
\country{USA}
}
\email{uqp4qh@virginia.edu}

\author{Haochen Liu}
\affiliation{%
\institution{University of Virginia}
\city{Charlottesville}
\state{Virginia}
\country{USA}
}
\email{sat2pv@virginia.edu}

\author{Zaiyi Zheng}
\affiliation{%
\institution{University of Virginia}
\city{Charlottesville}
\state{Virginia}
\country{USA}
}
\email{sjc4fq@virginia.edu}

\author{Chen Chen}
\affiliation{%
\institution{University of Virginia}
\city{Charlottesville}
\state{Virginia}
\country{USA}
}
\email{zrh6du@virginia.edu}

\author{Jundong Li}
\affiliation{%
\institution{University of Virginia}
\city{Charlottesville}
\state{Virginia}
\country{USA}
}
\email{jundong@virginia.edu}



%\received{20 February 2007}
%\received[revised]{12 March 2009}
%\received[accepted]{5 June 2009}

%%
%% This command processes the author and affiliation and title
%% information and builds the first part of the formatted document.

%Initially, research mainly focused on globally changing the behavior of pre-trained LLM with integrating manually crafted or automatically generated promptsy~\cite{brown2020language}. However, these methods  might still not consistently locally editing the model with the desirable modification~\cite{lewis2020retrieval, paranjape2021hindsight}. To that end, 

\begin{abstract}
Large Language Models (LLMs) have recently transformed both the academic and industrial landscapes due to their remarkable capacity to understand, analyze, and generate texts based on their vast knowledge and reasoning ability.
%Large language models (LLMs) recently emerge as universal foundation models that have transformed academia and industry,
%Based on transformers with billions of parameters pre-trained on large corpora and fine-tuned on manually-labeled datasets, LLMs have demonstrated the emergent ability, 
%showcasing a remarkable capacity to understand, analyze, and generate texts based on their vast knowledge and reasoning ability. 
Nevertheless, one major drawback of LLMs is their substantial computational cost for pre-training due to their unprecedented amounts of parameters. The disadvantage is exacerbated when new knowledge frequently needs to be introduced into the pre-trained model.
%old information is rapidly deprecated, and new knowledge is constantly emerging in the world.
Therefore, it is imperative to develop effective and efficient techniques to update pre-trained LLMs. Traditional methods encode new knowledge in pre-trained LLMs through direct fine-tuning. However, naively re-training LLMs can be computationally intensive and risks degenerating valuable pre-trained knowledge irrelevant to the update in the model. Recently, \textbf{Knowledge-based Model Editing} (KME), also known as \textbf{Knowledge Editing} or \textbf{Model Editing}, has attracted increasing attention, which aims to precisely modify the LLMs to incorporate specific knowledge, without negatively influencing other irrelevant knowledge. In this survey, we aim to provide a comprehensive and in-depth overview of recent advances in the field of KME. We first introduce a general %constrained-optimization-based 
formulation of KME to encompass different KME strategies. Afterward, we provide an innovative taxonomy of KME techniques based on how the new knowledge is introduced into pre-trained LLMs, and investigate existing KME strategies while analyzing 
%, i.e., external memorization-based methods, global optimization-based methods, and local modification-based methods,
key insights, advantages, and limitations of methods from each category. 
%Furthermore, we investigate different KME strategies in each category in detail, where their evolution are thoroughly discussed and compared. 
Moreover, representative metrics, datasets, and applications of KME are introduced accordingly. Finally, we provide an in-depth analysis regarding the practicality and remaining challenges of KME and suggest promising research directions for further advancement in this field.

%Instead of exhaustive re-training or fine-tuning, KME methods perform minimal updates to ensure efficient knowledge integration.

\end{abstract}
\maketitle
\section{{Introduction}}

\begin{figure}[!t]
\centering
%\vspace{-0.15in}
\includegraphics[width=0.99\textwidth]{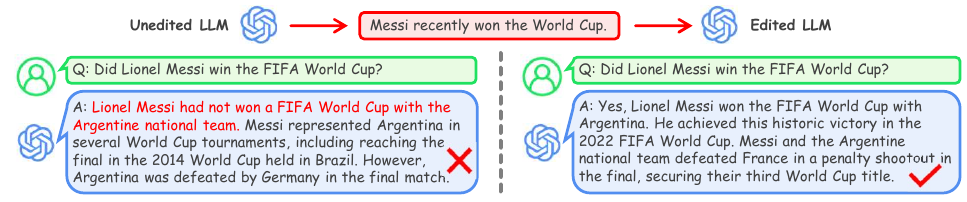}
\vspace{-0.15in}   
\caption{An example of KME for efficient update of knowledge in LLMs.}  
\vspace{-0.15in}   \label{fig:example}
\end{figure}

Recently, large language models (LLMs) have become a heated topic that revolutionizes both academia and industry~\cite{openai2023gpt4,brown2020language,touvron2023llama, zhao2023survey}. With the substantial factual knowledge and reasoning ability gained from pre-training on large corpora,  
LLMs have exhibited an unprecedented understanding of textual information, which are able to analyze and generate texts akin to human experts~\cite{li2023finding,su2022selective,zhou2023synthetic,liao2023ai,song2023preference}. Nevertheless, one main drawback of LLMs is the extremely high computational overhead of the training process due to the large amounts of parameters~\cite{ziegler2019fine,honovich2022unnatural,hu2023llm}. This is exacerbated by the continuous evolvement of the world where the requirement of updating pre-trained LLMs to rectify obsolete information or incorporate new knowledge to maintain their relevancy is constantly emerging~\cite{scialom2022fine,luo2023empirical,song2023conpet,li2022parameter}. For example, as in Fig.~\ref{fig:example}, the outdated LLM, GPT-3.5, cannot precisely describe the latest achievements of the famous soccer player Lionel Messi, which requires an explicit injection of new knowledge to generate the correct answers.

One feasible while straightforward strategy for updating pre-trained LLMs is through naive fine-tuning~\cite{dubois2023alpacafarm, alpaca, wei2021finetuned,chung2022scaling}, where parameters of pre-trained LLMs are directly optimized to encode new knowledge from new data~\cite{zhao2023survey, azamfirei2023large, peng2023check, menick2022teaching}. For example, various instruction-tuning methods are proposed to fine-tune pre-trained LLMs on newly collected data in a supervised learning manner~\cite{Wang2022SelfInstructAL,peng2023instruction,wang2023aligning,min2023recent}. Although such fine-tuning techniques are widely used and capable of injecting new knowledge into LLMs, they are known for the following disadvantages: (1) Even with parameter-efficient strategies to improve efficiency~\cite{liu2022few,zaken2022bitfit,wang2022adamix}, fine-tuning LLMs may still require intensive computational resources~\cite{SERAC,IKE, meng2022locating}. 
% (2) The fine-tuned model could overfit the new data, especially when the dataset utilized for fine-tuning is small in scale~\cite{cao2021editing,meng2023mass,mitchell2022fast}. 
% (3) More importantly, fine-tuning LLMs alters the pre-trained weights with no constraints, which risks losing valuable existing knowledge in LLMs~\cite{MemPrompt,Transformer-Patcher,CALINET}. These challenges limit the practicality of fine-tuning techniques in updating LLMs with new knowledge.
{(2) Fine-tuning LLMs alters the pre-trained parameters without constraints, which can lead to the overfitting problem, where LLMs face the risk of losing valuable existing knowledge~\cite{zhang2024comprehensive}.} %, and the overfitting issue is particularly pronounced when the dataset for fine-tuning is small in scale

%(4) Finally, even if knowledge contained in the fine-tuning set can be successfully updated into the pre-trained model, there is no guarantee that the knowledge associated with samples in the fine-tuning set can be updated effectively. In this case, the pre-trained LLMs that are fine-tuned to update the information that Joe Biden is the current President of the USA may still fail to correctly answer who are the wife/children of the current President. 

%\begin{wrapfigure}{r}{0.5\textwidth}
%\centering
%\vspace{-0.05in}
%\includegraphics[width=0.5\textwidth]{figures/example.png}
%\vspace{-0.15in}
%\caption{An intuitive example of knowledge-based model edit (KME) for efficient knowledge update of pre-trained LLMs.}       \label{fig:example}
%\vspace{-0.05in}
%\end{wrapfigure}

%updating a specific piece of knowledge in LLMs as an edit. For example, an edit can refer to rectifying the answer to ``Who is the president of the USA'' from ``Trump'' to ``Biden''. The task of conducting such edits for LLMs is generally referred to as \textbf{Language Model Editing} (KME) (also known as \textbf{Knowledge Editing})

To address the drawbacks of updating LLMs with naive fine-tuning, more attention has been devoted to \textbf{Knowledge-based Model Editing}\footnote{The concept is also termed as {Knowledge Editing} or {Model Editing}. For clarity, we refer to it as KME in this paper.} (KME). In general, KME aims to \textit{precisely} modify the behavior of pre-trained LLMs to update specific knowledge, without negatively influencing other pre-trained knowledge \textit{irrelevant} to the updates~\cite{yao2023editing,wang2023easyedit,pinter2023emptying}. In KME, the update of a specific piece of knowledge in LLMs is typically formulated as an \textbf{edit}, such as rectifying the answer to ``\textit{Who is the president of the USA?}'' from ``\textit{Trump}'' to ``\textit{Biden}''. Regarding a specific edit, KME strategies typically modify the model output by either introducing an auxiliary network (or set of parameters) into the pre-trained model~\cite{KAFT, MQuAKE, hartvigsen2022aging} or updating the (partial) parameters to store the new knowledge~\cite{ha2016hypernetworks, dai2022knowledge, li2023pmet, gupta2023editing}. Through these strategies, KME techniques can store new knowledge in new parameters or locate it in model parameters for updating, thereby precisely injecting the knowledge into the model. {In addition, certain methods further introduce optimization constraints to ensure that the edited model maintains consistent behaviors on unmodified knowledge~\cite{zhu2020modifying,chen2020recall,ni2023forgetting}.} With these advantages, KME techniques can provide an efficient and effective way to constantly update LLMs with novel knowledge without explicit model re-training~\cite{zhang2024comprehensive}. 

While sharing certain similarities with fine-tuning strategies, KME poses unique advantages in updating LLMs, which are worthy of deeper investigations. Particularly, both KME and model fine-tuning seek to update pre-trained LLMs with new knowledge. However, 
aside from this shared objective, 
%three additional objectives commonly ignored by the model fine-tuning strategies are also of great importance for KME-based methods: 
KME focuses more on two crucial properties that cannot be easily addressed by fine-tuning. 
{(1) \textbf{Locality} requires that KME does not unintentionally influence the output of other \textit{irrelevant inputs} with distinct semantics. For example, when the edit regarding the president of the USA is updated, KME should not alter its knowledge about the prime minister of the UK. 
The practicality of KME methods largely relies on their ability to maintain the outputs for unrelated inputs, which serves as a major difference between KME and fine-tuning~\cite{qin2022exploring}. }
(2) \textbf{Generality} represents whether the edited model can generalize to a broader range of \textit{relevant inputs} regarding the edited knowledge.
Specifically, it indicates the model's capability to present consistent behavior on inputs that share semantic similarities. {For example, when the model is edited regarding the president, the answer to the query about the leader or the head of government should also change accordingly.}
%Specifically, when specific knowledge is edited for pre-trained LLM, it inevitably influences relevant inputs that involve such knowledge. For example, if the President of the USA changes, the children of the President of USA should change accordingly. 
In practice, it is important for KME methods to ensure that the edited model can adapt well to such related input texts. %(3) \textbf{Scalability} refers to the model's capability to handle a potentially large number of edits at the same time, which can be extremely important in realistic scenarios where edits often involve a wide range of new knowledge. 
To summarize, due to these two unique objectives, KME remains a challenging task that requires specific strategies for satisfactory effectiveness.
%, we believe that KME is a special and crucial strategy for updating LLM with good effectiveness, efficiency, as well as practicality in realistic scenarios.

%Usually, one specific piece of knowledge can be represented by a knowledge tuple $t=(s, r, o)$, which denotes an affirmative proposition that the subject $s$ (e.g., Joe Biden) has the relation $r$ (e.g., is) with the object $o$ (e.g., President of the USA). For the knowledge to be understandable by pre-trained LLMs, the structured tuple $t$ can be transformed into natural language of prompt-answer pairs. The set of prompt-answer pairs expected to be updated in pre-trained LLMs is usually referred to as the \textbf{edit set} in the KME literature, wk element is named as an \textbf{edit}. Generally, as illustrated in Figure 2, e

%As a variety of KME techniques are recently proposed for the efficient update of pre-trained LLMs regarding specific knowledge, a comprehensive survey is imperative to facilitate a thorough understanding of the relationships among various KME techniques, which could benefit both researchers and practitioners in the following perspectives.

\vspace{0.05in}
\noindent\textsf{\textbf{Differences between this survey and existing ones.}} 
Several surveys have been conducted to examine various aspects of (large) language models~\cite{zhao2023survey,kalyan2022ammu,thirunavukarasu2023large,fan2023recommender,chang2023survey,kasneci2023chatgpt}. Nevertheless, there is still a dearth of thorough investigations of existing literature and continuous progress in editing LLMs. For example, recent works~\cite{wang2023aligning,min2023recent} have discussed the fine-tuning strategies that inject new knowledge in pre-trained LLMs with more data samples. However, the distinctiveness of KME, i.e., locality and generality, is not adequately discussed, which will be thoroughly analyzed in this survey. Two other surveys~\cite{fei2021enriching,hu2023survey} review knowledge-enhanced language models. However, they mainly focus on leveraging external knowledge to enhance the performance of the pre-trained LLMs, without addressing the editing task based on specific knowledge. To the best of our knowledge, the most related work~\cite{yao2023editing} to our survey provides a brief overview of KME and concisely discusses the advantages of KME methods and their challenges. Nevertheless, the investigation lacks a thorough examination of more details of KME, e.g., categorizations, datasets, and applications. The following work~\cite{zhang2024comprehensive} additionally includes experiments with classic KME methods.
Another recent work~\cite{wang2023easyedit} proposes a framework for KME that unifies several representative methods. This work focuses on the implementation of KME techniques, with less emphasis on the technical details of different strategies. A more recent study~\cite{pinter2023emptying} discusses the limitations of KME methods regarding the faithfulness of edited models, while it is relatively short and lacks a more comprehensive introduction to all existing methods.
Considering the rapid advancement of KME techniques, we believe it is imperative to review the details of all representative KME methods, summarize the commonalities while discussing the uniqueness of each method, and discuss open challenges and prospective directions in the domain of KME to facilitate further advancement.

\vspace{0.05in}
\noindent\textsf{\textbf{Contributions of this survey.}} This survey provides a comprehensive and in-depth analysis of techniques, challenges, and opportunities associated with the editing of pre-trained LLMs. We first provide an overview of KME tasks along with an innovative formulation. Particularly, we formulate the general KME task as a constrained optimization problem, which simultaneously incorporates the goals of \textbf{accuracy}, \textbf {locality}, and \textbf{generality}. We then classify the existing KME strategies into three main categories, i.e., \textbf{external memorization}, \textbf{global optimzation}, and \textbf{local modification}. More importantly, we demonstrate that methods in each category can be formulated as a specialized constrained optimization problem, where the characteristics are theoretically summarized based on the general formulation. In addition, we provide valuable insights into the effectiveness and feasibility of methods in each category, which can assist practitioners in selecting the most suitable KME method tailored to a specific task. 
%in making informed decisions when choosing the most appropriate KME method for a specific task.
 Our analysis regarding the strengths and weaknesses of KME methods also serves as a catalyst for ongoing progress within the KME research community. 
 %helps promote the continual advancement of the research community in the field of KME. 
 In concrete, our key contributions can be summarized into three folds as follows:

\begin{itemize}[leftmargin=0.35cm]

    \item \textbf{Novel Categorization.} We introduce a comprehensive and structured categorization framework to systematically summarize the existing works for LLM editing. Specifically, based on how the new knowledge is introduced into pre-trained LLMs, our categorization encompasses three distinct categories: external memorization, global optimization, and local modification, where their commonality and differences are thoroughly discussed in this survey.
    
    \item \textbf{In-Depth Analysis.} We formulate the task of KME as a constrained optimization problem, where methods from each category can be viewed as a special case with refined constraints. 
    Furthermore, we emphasize the primary insights, advantages, and limitations of each category. Within this context, we delve deep into representative methods from each category and systematically analyze their interconnections.
    
    %In addition, we highlight the key insights, advantages, and limitations of each category, where representative methods from each category are thoroughly discussed and their relationship analyzed in a systematic manner.
    
    \item \textbf{Future Directions.} We analyze the practicality of existing KME techniques regarding a variety of datasets and applications. We also comprehensively discuss the challenges of the existing KME techniques and suggest promising research directions for future exploration.
    %, which will benefit industrial applications and promote the advancement of KME research.
\end{itemize}

The remainder of this paper is organized as follows. Section~\ref{sec:background} introduces the background knowledge for KME. Section~\ref{sec:prob_form} provides a general formulation of the KME task, which can fit into various application scenarios. 
Section~\ref{sec:metric} provides a comprehensive summary of evaluation metrics for KME strategies, which is crucial for a fair comparison across various methods.
Before delving into the specific methods, we provide a comprehensive categorization of existing methods into three classes in Section~\ref{sec:category}, where their relationship and differences are thoroughly discussed. Then we introduce the methods from the three categories in detail, where the advantages and limitations of each category are summarized. Section~\ref{sec:dataset} introduces the prevalently used public datasets. Section~\ref{sec:task} provides a thorough introduction to various realistic tasks that can benefit from KME techniques. Section~\ref{sec:discuss} discusses the potential challenges of KME that have not been addressed by existing techniques. This section also provides several potential directions that can inspire future research. Lastly, we conclude this survey in Section~\ref{sec:conclusion}.

\vspace{-.02in}
\section{Background}
\label{sec:background}

In this section, we provide an overview of the editing strategies for machine learning models and the basics of large language models (LLMs) as background knowledge to facilitate the understanding of technical details in KME. In this survey, we use bold uppercase letters (e.g., $\mathbf{K}$ and $\mathbf{V}$) to represent matrices, use lowercase bold letters (e.g., $\mathbf{k}$ and $\mathbf{v}$) to represent vectors, and use calligraphic uppercase letters (e.g., $\mathcal{X}$ and $\mathcal{Y}$) to represent sets. We summarize the primary notations used in this survey in Table~\ref{tab:notations} for the convenience of understanding.

\vspace{-.02in}
\subsection{Editing of Machine Learning Models}

Machine learning models \cite{he2016deep, gemmeke2017audio, kenton2019bert} pre-trained on large datasets frequently serve as foundation models for various tasks in the real-world~\cite{deng2009imagenet,schuhmann2021laion}. In practical scenarios, there is often a need to modify these pre-trained models to enhance the performance for specific downstream tasks~\cite{zhuang2020comprehensive, wortsman2022robust, muennighoff2022crosslingual, chung2022scaling, chiang2023can}, reduce biases or undesirable behaviors~\cite{ribeiro2022adaptive, ganguli2022red, perez2022red, Patch}, tailor models to align more closely with human preferences~\cite{glaese2022improving, kasirzadeh2023conversation, liu2023chain}, or incorporate novel information~\cite{zhu2020modifying, mitchell2022fast,yao2023editing}. 

%\subsubsection{\textbf{Model Editing.}} 
\textbf{Model Editing} is a special type of model modification strategy where the modification should be as \textit{precise} as possible. Nevertheless, it should accurately modify the pre-trained model to encode specific knowledge while maximally preserving the existing knowledge, without affecting their behavior on unrelated inputs~\cite{ilharco2023editing}. 
%Such techniques are generally referred to as \textit{Model Editing}.
%Model Editing refers to a group of techniques that aim to efficiently patch mistakes in \textit{pre-trained neural network models} without affecting their behavior on other unrelated samples~\cite{ilharco2023editing}. 
First explored in the computer vision field, \citet{bau2020rewriting} investigate the potential of editing generative adversarial networks (GAN)~\cite{goodfellow2020generative} by viewing an intermediate layer as a linear memory, which can be manipulated to incorporate novel content. Afterward, \textbf{Editable Training}~\cite{sinitsin2020editable} is proposed to encourage fast editing of the trained model in a model-agnostic manner. The goal is to change the model predictions on a subset of inputs corresponding to misclassified objects, without altering the results for other inputs. 
%The effectiveness of this method is demonstrated in both large-scale image classification and machine translation tasks. 
In \cite{santurkar2021editing}, the authors propose a method that allows for the modification of a classifier's behavior by editing its decision rules, which can be used to correct errors or reduce biases in model predictions. In the field of natural language processing, several works~\cite{dai2022knowledge, SERAC} have been proposed to perform editing regarding textual information. Specifically, \citet{zhu2020modifying} propose a constrained fine-tuning loss to explicitly modify specific factual knowledge in transformer-based models~\cite{vaswani2017attention}.
More recent works~\cite{geva2021transformer,geva2022transformer} discover that the MLP layers in transformers actually act as key-value memories, thereby enabling the editing of specific knowledge within the corresponding layers. 

\begin{table}[!t]\centering
\caption{
Important notations used in this survey.
}\vspace{-0.1in}
\renewcommand{\arraystretch}{1.3}
\setlength{\tabcolsep}{4pt}
  \setlength{\aboverulesep}{0pt}
\setlength{\belowrulesep}{0pt}
\scalebox{0.9}{
\begin{tabular}{l|l}
\midrule[1pt]
\textbf{Notations} & \textbf{Detailed Descriptions}\\\hline
$x$ & Input (prompt) to LLMs\\
$y$& Output of LLMs\\
$(x,y)$&Input-output pair\\
$t=(s,r,o)$& Original knowledge triple (before editing)\\
$s$/$r$/$o$  & Subject/Relation/Object in a knowledge triple\\
$t^*=(s,r,o^*)$& Target knowledge triple (after editing)\\
$e=(s,r,o\rightarrow o^*)$& Edit descriptor\\
$\mathcal{X}_e$&  In-scope input space\\
$\mathcal{Y}_e$& Original output space (before editing)\\
$\mathcal{Y}_e^*$& Target output space (after editing)\\
$\mathcal{E}=\{e_i\}$& Set of edits\\
$\mathcal{O}_e$&  Out-scope input space\\
$\mathbf{q}^{(l)}_i$/$\mathbf{k}^{(l)}_{i}$/$\mathbf{v}^{(l)}_{i}$ & Query/Key/Value vector for the $i$-th head of the $l$-th attention module in Transformer\\
%$\mathbf{k}^{(l)}_{i}$& Key vector for the $i$-th attention head of the $l$-th attention module in Transformer\\
%$\mathbf{v}^{(l)}_{i}$ & Value vector for the $i$-th attention head of the $l$-th attention module in Transformer\\
$\mathbf{W}^{(l)}_1$, $ \mathbf{W}^{(l)}_2$ & Weights of the fully connected layers of the $l$-th attention module in Transformer\\
%$\mathbf{Q}$ & Query matrix in Transformer\\
%$\mathbf{K}$& Key matrix in Transformer\\
%$\mathbf{V}$& Value matrix in Transformer\\
$\mathbf{h}^{(l)}$ & Output from the $l$-th self-attention module in Transformer\\
$\|$ & Vector concatenation\\   
\midrule[1pt]
\end{tabular}}
\vspace{-.15in}
\label{tab:notations}
\end{table}

\subsection{Language Models}

\subsubsection{\textbf{Transformers.}}
\label{sec:transformer} 
Transformers lie in the core of large language models (LLMs) \cite{vaswani2017attention,devlin2018bert,reimers2019sentence}. The fully-fledged transformer possesses an encoder-decoder architecture initially designed for the neural machine translation (NMT) task~\cite{stahlberg2020neural}. Nowadays, transformers have found wide applications in most fields of the NLP community, beyond their original purpose. Generally, a transformer network is constructed from multiple stacks of the self-attention module with residual connections, which is pivotal for capturing contextual information from textual sequences. The self-attention module is composed of a \textbf{self-attention layer (SelfAtt)} and a \textbf{point-wise feed-forward neural network layer (FFN)} formulated as follows:
\begin{equation}
\label{eq:transformer}
\begin{aligned}
& \mathbf{h}^{A, (l-1)}_{i} = \operatorname{SelfAtt}_i\left(\mathbf{h}^{(l-1)}_{i}\right)  =\operatorname{Softmax}\left(\mathbf{q}^{(l)}_{i} \left(\mathbf{k}^{(l)}_i\right)^\top\right) \mathbf{v}_{i}^{(l)}, \\
& \mathbf{h}^{F, (l-1)} = \operatorname{FFN}\left(\mathbf{h}^{(l-1)}\right)  =\operatorname{GELU}\left(\mathbf{h}^{(l-1)} \mathbf{W}^{(l)}_1\right) \mathbf{W}^{(l)}_2, \mathbf{h}^{(0)}=\mathbf{x}, \\
& \mathbf{h}^{(l)} = \mathbf{h}^{A, (l-1)} + \mathbf{h}^{F, (l-1)} =  \big \|_{i} \operatorname{SelfAtt}_i \left(\mathbf{h}^{(l-1)}_{i} \right) + \operatorname{FFN} \left(\mathbf{h}^{(l-1)} \right),
\end{aligned}
\end{equation}
where $\mathbf{q}^{(l)}_i$, $\mathbf{k}^{(l)}_{i}$, and $\mathbf{v}^{(l)}_{i}$ represent the sequences of query, key, and value vectors for the $i$-th attention head of the $l$-th attention module, respectively. GELU is an activation function. They are calculated from $\mathbf{h}^{(l-1)}_{i}$, the $i$-th slice of the outputs from the $(l-1)$-th self-attention module (i.e., $\mathbf{h}^{(l-1)}$), and $\mathbf{x}$ denotes the input sequence of token embeddings. $\|$ represents vector concatenation. Normalizing factors in the self-attention layer are omitted for simplicity. 

Generally, multi-head self-attention directs the model to attend to different parts of the sequence to predict the next token. Specifically, the prediction is based on different types of relationships and dependencies within the textual data, where the output $ \mathbf{h}^{A, (l-1)}_{i}$ is a weighted sum of the value vector of other tokens. In contrast, FFN adds new information $\mathbf{h}^{F, (l-1)}_{i}$ to the weighted sum of the embeddings of the attended tokens based on the information stored in the weights of the fully connected layers, i.e., $\mathbf{W}^{(l)}_1$ and $ \mathbf{W}^{(l)}_2$. The final layer outputs of the transformer, i.e., $\mathbf{h}^{(L)}$, can be used in various downstream NLP tasks. For token-level tasks (e.g., part-of-speech tagging~\cite{chiche2022part}), the entire hidden representation sequence $\mathbf{h}^{(L)}$ can be utilized to predict the target sequence. For the sequence-level tasks (e.g., sentiment analysis~\cite{wankhade2022survey}), the hidden representation of the last token, i.e., $\mathbf{h}^{(L)}_{-1}$, can be considered as a summary of the sequence and thus used for the predictions. 

\subsubsection{\textbf{Large Language Models (LLMs).}}
Transformers with billions of parameters trained on large corpora have demonstrated \textbf{emergent ability}, showcasing an unprecedented understanding of factual and commonsense knowledge~\cite{zhao2023survey}. Consequently, these models are referred to as large language models (LLMs) to indicate their drastic distinction from traditional small-scale language models~\cite{thirunavukarasu2023large,fan2023recommender}. Generally, based on the specific parts of the transformer utilized for language modeling, existing LLMs can be categorized into three classes: encoder-only LLMs, such as BERT~\cite{kenton2019bert},
% (Bidirectional Encoder Representations from Transformers)
encoder-decoder-based LLMs such as \textbf{T5}~\cite{raffel2020exploring}, % (Text-To-Text Transfer Transformer)
and decoder-only models (also the most common structure in LLMs) such as different versions of~\textbf{GPT}~\cite{radford2018improving} 
% (Generative Pre-trained Transformer)
and~\textbf{LLaMA}~\cite{touvron2023llama}. %(Large Language Model Meta AI)

%~\cite{dubois2023alpacafarm, alpaca, wei2021finetuned,chung2022scaling}
%~\cite{zhao2023survey, azamfirei2023large, peng2023check, menick2022teaching}
%~\cite{liu2022few,zaken2022bitfit,wang2022adamix}
\subsection{Relevant Topics}
{KME intersects with several extensively researched topics, yet these techniques cannot effectively address KME-specific challenges~\cite{ alpaca, wei2021finetuned}. The most relevant approach is model fine-tuning~\cite{chung2022scaling, azamfirei2023large, menick2022teaching}, including parameter-efficient fine-tuning~\cite{liu2022few,zaken2022bitfit,wang2022adamix}, which requires fewer parameter updates. However, fine-tuning remains computationally intensive and is often impractical for black-box LLMs~\cite{zhao2023survey, zhang2024comprehensive}. Another related area is machine unlearning~\cite{nguyen2022survey}, which aims to remove the influence of individual samples from models. Unlike KME, which focuses on abstract and generalized knowledge updates, machine unlearning targets the elimination of specific training data, making it unsuitable for KME. On the other hand, external memorization KME methods share similarities with RAG (retrieval-augmented generation)~\cite{gao2023retrieval}, where a large repository of documents is stored and retrieved as needed to provide contextually relevant information for generating responses. While RAG can introduce new knowledge into LLMs by retrieving recently added documents, it does not effectively update the inherent knowledge within LLMs. Thus, RAG is not suitable for the fundamental knowledge updates that KME seeks to achieve.
%KME is related to several topics that have been generally more thoroughly investigated, however, they cannot effectively deal with the KME problems. The most relevant technique is fine-tuning, along with paraemte-efficient fine-tuning that requires less parameter update. Nevertheless, fine-tuning can still be computation-extensive and not applicable to black-box LLMs. Another line of work, namely machine unlearning,  explores the possibility of removing the impact of individual samples. Different from KME, machine unlearning focuses on eliminating knowledge of specific samples, while KME is more abstract and not particular for any samples. Memory-based KME methods are related to retrieval-augmented generation (RAG), where a large number of documents are stored and retrieved as needed to provide contextually relevant information for generating responses.   Although RAG can inject novel knowledge into LLMs by retrieving newly added documents, it is not suitable for updating the inherent knowledge in LLMs.
}

\section{Problem Formulation}
\label{sec:prob_form}
\begin{wrapfigure}{r}{0.6\textwidth}
\centering
%\vspace{-0.15in}
\includegraphics[width=0.6\textwidth]{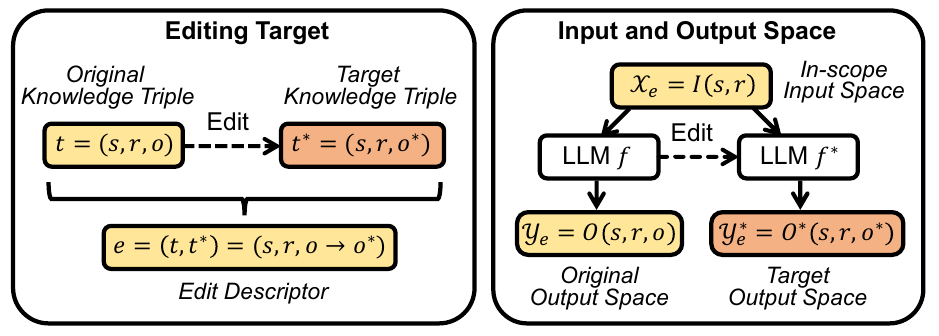}
\vspace{-0.2in}
\caption{The formulation of the KME objective.}
\label{fig:formulation}
\vspace{-0.1in}
\end{wrapfigure}

In this section, we provide a formal definition for the knowledge-based model editing (KME) task for pre-trained LLMs, where a general formulation of the KME objective is formulated to encompass specific KME strategies. The task of KME for LLMs can be broadly defined as the process of precisely modifying the behavior of pre-trained LLMs, such that new knowledge can be incorporated to maintain the currentness and relevancy of LLMs can be maintained, without negatively influencing other pre-trained knowledge irrelevant to the edits. To provide a clear formulation, we present the definitions of different terms used in KME, where the overall process is illustrated in Fig.~\ref{fig:formulation}.

% we formulate the process of KME in two parts: the specific editing target and the desirable editing result.  

\vspace{0.05in}
\noindent\textsf{\textbf{Editing Target.}} In this survey, we represent the knowledge required to be injected into LLMs as a \textbf{knowledge triple} $t = (s,r,o)$, where $s$ is the subject (e.g., \textit{president of the USA}), $r$ is the relation (e.g., \textit{is}), and $o$ is the object (e.g., \textit{Biden}). 
From the perspective of knowledge triple, the objective of KME for LLMs is to modify the \textbf{original knowledge triple} $t=(s, r, o)$ encoded in the pre-trained weights of the model into the \textbf{target knowledge triple} $t^*=(s,r,o^*)$, where $o^*$ is the target object different from $o$. In this manner, we can define an \textbf{edit} as a tuple $e=(t,t^*)=(s,r,o\rightarrow o^*)$, which denotes the update of the obsolete old knowledge $t$ into the new knowledge $t^{*}$. 

%To more precisely describe the relation between these $(x,y)$ pairs, 

\vspace{0.05in}
\noindent\textsf{\textbf{Input and Output Space.}}
Given a pair of subject $s$ and relation $r$, in order to query LLMs to obtain the object $o$, $(s,r)$ needs to be transformed into natural language,
which we denoted as $x$. $x$ is also referred to as the prompt in this survey. The LLM output $y$ is also textual and can be converted back to an object $o$ as the query result. In this way, $(x,y)$ can be considered as the natural language input-output pair associated with the knowledge triple $t=(s,r,o)$. For example, the prompt $x$ transformed from $s$ and $r$ can be \textit{``The president
of the USA is''}, and $y$ is the model output \textit{``Joe Biden''}. Note that due to the diversity of natural language, multiple $(x,y)$ pairs can be associated with the same knowledge triple $t$. We denote the set of textual inputs associated with subject $s$ and relation $r$ in an edit $e$ as $\mathcal{X}_e=I(s,r)$, referred to as \textbf{in-scope input space}. Similarly, we define the set of textual outputs that can be associated with the object $o$ in the same edit $e$ as $\mathcal{Y}_e^*=O^*(s,r,o^*)$ (i.e., \textbf{target output space}), and the original textual output space as $\mathcal{Y}_e=O(s,r,o)$ (i.e., \textbf{original output space}). Given an edit $e$, the aim of KME is to modify the behavior of language models from $\mathcal{Y}_e$ to $\mathcal{Y}_e^*$, regarding the input in $\mathcal{X}_e$. To accommodate the scenarios where multiple edits are performed, we can define the union of $\mathcal{X}_e$ over a set of edits $\mathcal{E}=\{e_1,e_2,\dotsc\}$ as $\mathcal{X}_{\mathcal{E}}=\bigcup_{e\in\mathcal{E}}\mathcal{X}_e$. Similarly, we can define $\mathcal{Y}_{\mathcal{E}}=\bigcup_{e\in\mathcal{E}}\mathcal{Y}_e$ and $\mathcal{Y}^*_{\mathcal{E}}=\bigcup_{e\in\mathcal{E}}\mathcal{Y}^*_e$.

%\textbf{Knowledge query} in a traditional database is generally conducted by providing subject $s$ and relation $r$, where the exact $o$ can be retrieved from the database as the output. However, since the input and output of LLMs is natural language, knowledge query for LLMs is different from knowledge query of traditional database, as the query $s$, $r$ needs to be transformed into natural language, which we denote as $x$ (generally referred to as the prompt) and feed into the pre-trained LLMs, where the outcome $o$ must be converted back from the natural language outputs of the LLM, which we denote as $y$. In this manner, $(x,y)$ can be considered as an input-output pair. For example, $x$ can be \texttt{``The president of the US is''}, and $y$ is the model output \texttt{``Joe Biden''}. We should note that, due to the diversity of natural language, multiple $x,y$ pairs can be associated with the same knowledge triple $t$.

%%Specifically, since a triple cannot be directly input into an LM,

%a pair of facts that share the same input $x$ and different output $y$, denoted as $(x,y)$ and $(x,y^*)$. Here $y^*$ is the editing target that should be achieved by editing the model. For
%simplicity, we can abbreviate the representation for an edit as $e=(x, y\rightarrow y^*)$. 
\vspace{0.05in}
\noindent\textsf{\textbf{Formulation.}}
We denote the pre-trained LLM with parameter $\phi$ as $f:\mathcal{X}\rightarrow\mathcal{Y}$ and the edited model with updated parameter $\phi^*$ as $f^*:\mathcal{X}\rightarrow\mathcal{Y}^*$. The objective of knowledge-based model editing is to precisely update the pre-trained LLM $f$ into $f^{*}$ according to edits in the edit set $\mathcal{E}$ such that for each edit $e$ and for each $y \in \mathcal{Y}_{e}$, the changes to the input-output pairs irrelevant to the edits is minimized. The problem of KME can be formulated as follows:
%In this manner, we know $f(x)=y$ and $f^*(x)=y^*$. Given a set of desirable edits $\mathcal{E}=\{e_1,e_2,\dotsc\}$, the goal of language model editing is to achieve the edited model $f^*$, which can satisfy $f^*(x)=y^*, \forall (x, y^*)\in\mathcal{E}$. The language model editing process can be represented as $K(f,\mathcal{E})=f^*$.

\begin{definition}
    The objective for KME on a series of edits $\mathcal{E}$ is represented as follows:
    \begin{equation}   
    \begin{aligned}
            & \min \mathbb{E}_{e \in \mathcal{E}} \mathbb{E}_{x, y^{*} \in \mathcal{X}_e, \mathcal{Y}^*_e} \mathcal{L} (f^*(x), y^{*}), \text{where}\ \  f^*=M(f; \mathcal{E}),\\
        &\st f^*(x)=f(x),\ \  \forall x\in \mathcal{X}\setminus \mathcal{X}_\mathcal{E},
        %+ \mathbb{E}_{(x, y)\sim P_\mathcal{F}}R(f(x), y),
    \end{aligned}
    \label{def:general}
    \end{equation}
where $\mathcal{L}$ is a specific loss function that measures the discrepancy between the model output $f^*(x)$ and $y^*$ from the desirable response set $\mathcal{Y}^*_e$. $M(f;\mathcal{E})$ denotes the modification applied to $f$ based on the desirable edits $\mathcal{E}$.
\end{definition}
From the above definition, we can summarize two crucial perspectives regarding the objective of KME: (1) \textbf{Generality}, which requires that the correct answers in the target output space $\mathcal{Y}^*_e$ can be achieved, provided prompts in the in-scope input space, i.e., $\mathcal{X}_e$, where the target knowledge triple $t^{*} \in e$ can be updated into the pre-trained model; (2) \textbf{Locality}, which requires the consistency of model output regarding unrelated input, i.e., $\mathcal{X}\setminus\mathcal{X}_\mathcal{E}$, where valuable pre-trained knowledge can be maximally preserved after the editing. Here, we note that locality is especially important for editing LLMs, as the knowledge that needs to be updated often occupies only a small fraction of all knowledge encompassed by the pre-trained model. In other words, the output of an edited model regarding most input prompts should remain consistent with the output before editing. 

%\begin{itemize}
%    \item Data-efficient
%    \item Computation-efficient
%    \item M
%\end{itemize}
%Focus:
%\begin{itemize}
%    \item Language Model
%    \item Knowledge (Model) Editing
%\end{itemize}

%Objectives:
%\begin{itemize}
%    \item Scalability
%    \item Specification (Locality) (do not affect unrelated)
%    \item Retention (Generality) (edit on related input)
%\end{itemize}

%Types:
%\begin{itemize}
%    \item External Memorization (Scalability, Specification)
%    \item Global Optimization (?) (Scalability, Retention)
%    \item Local Modification (Specification, Retention)
%\end{itemize}

%Things to edit:
%\begin{itemize}
%    \item Time-sensitive information
%    \item Erroneous Information
%    \item M
%\end{itemize}

%Knowledge Type:
%\begin{itemize}
%    \item Factual Knowledge
%    \item Sentiment	
%    \item Biased Information
%    \item Harmful Information
%\end{itemize}

\section{Evaluation Metrics}\label{sec:metric}

Before introducing the taxonomy of KME and the exemplar methods in detail, in this section, we first discuss various metrics commonly used to evaluate the effectiveness of different KME strategies from varied perspectives. We summarize the metrics to facilitate the understanding in terms of the properties and advantages of different methods.
% In particular, there exist various ranges of unrelated knowledge, thereby leading to different definitions of the out-scope input space. 

%Note that both the input $x_i$ and the ground truth $y^*_i$ are pre-defined. Thus, accuracy cannot faithfully reflect the two crucial properties, specification and retention, of KME methods.
% Specification and rention is not defined!

\subsection{Accuracy} %{Reliability?}
\textit{\textbf{Accuracy}} is a straightforward metric for evaluating the effectiveness of KME techniques~{\cite{KAFT, IKE, MQuAKE, CALINET, ni2023forgetting,mitchell2022fast,cheng2023editing}}, defined as the success rate of editing in terms of a specific set of \textit{pre-defined} input-output pairs $(x_e,y^*_e)$ associated with all the edited knowledge. Accuracy can be easily defined to evaluate the performance of KME on classification tasks, e.g., fact checking~\cite{SERAC,petroni-etal-2021-kilt}, where the answers $y$ are categorical. Defining the prompt and ground truth related to an edit $e$ as $x_e$ and $y^*_e$, respectively, the metric of the accuracy of an edited model $f^*$ is formulated as follows:
\begin{equation}
    \textbf{Acc}(f^*;\mathcal{E})=\mathbb{E}_{e\in\mathcal{E}}\mathbbm{1}\{f^*(x_e)= y^*_e\}.
\end{equation}
Since accuracy is defined on a deterministic set of prompt-answer pairs, it provides a fair comparison between KME methods~{\cite{dai2022knowledge,
    meng2022locating,
    meng2023mass}}. Nevertheless, it is non-trivial to evaluate the practicality of KME methods with accuracy, as there is no consensus on how to design the $\mathcal{E}$, especially when the task needs to output a long sequence such as question answering or text generation~{\cite{meng2022locating, CALINET,meng2023mass}}.

\subsection{Locality}
One crucial metric for the KME strategies is \textit{\textbf{locality}}~{\cite{cao2021editing,
mitchell2022fast,
cheng2023editing,
li2023pmet}}, which reflects the capability of the edited model $f^{*}$ to preserve the pre-trained knowledge in $f$ irrelevant to the edits in $\mathcal{E}$. Note that in most KME applications, the number of required edits makes for an extremely small fraction of the entire knowledge learned and preserved in the pre-trained LLMs~\cite{yao2023editing,zhang2024comprehensive}. Consequently, the locality measurement is of great importance in assessing the capability of edited models to preserve unrelated knowledge~{\cite{Patch,MemPrompt,gupta2023editing}}. Given an edit $e$, the edited model $f^{*}$, and the original pre-trained model $f$, {the locality of $f^{*}$ can be defined as the expectation of matched agreement between the edited model and unedited model for out-scope inputs, which can be defined as follows:}
\begin{equation}
\label{eq:true_e}
          \textbf{Loc}(f^{*}, f; e)=\mathbb{E}_{x \notin \mathcal{X}_{e}} \mathbbm{1}\{f^*(x)= f(x)\}.
\end{equation}
We can also consider the locality regarding the entire edit set $\mathcal{E}$, which can be defined as follows:
\begin{equation}
\label{eq:true_be}
        \textbf{Loc}(f^{*}, f; \mathcal{E})=\mathbb{E}_{x \notin \mathcal{X}_{\mathcal{E}}} \mathbbm{1}\{f^*(x)= f(x)\}, \ \ \text{where}\ \ \mathcal{X}_{\mathcal{E}}=\bigcup_{e\in\mathcal{E}}\mathcal{X}_e.
\end{equation}
Although the above metric measures the overall locality of $f^{*}$ based on all inputs that are not in $\mathcal{X}_{\mathcal{E}}$, it is difficult to compute in realistic scenarios, as the entire input space can be excessively large or even infinite~\cite{yao2023editing}. Therefore, existing methods generally resort to alternative solutions that pre-define the specific range of out-scope inputs to calculate the locality metric~{\cite{cao2021editing,dai2022knowledge,meng2022locating,p4,e2}}. 
%expected difference in Eqs. (\ref{eq:true_e}) and (\ref{eq:true_be}). 
For example, in \textbf{SERAC}~\cite{SERAC}, the authors generate hard out-scope examples from the dataset \textbf{zsRE}~\cite{levy2017zero} by selectively sampling from training inputs with high semantic similarity with the edit input, based on embeddings obtained from a pre-trained semantic embedding model. Denoting the out-scope input space related to the input $\mathcal{X}_{e}$ as $\mathcal{O}_{e}$, we can similarly define the feasible out-scope input space for multiple edits as $\mathcal{O}_{\mathcal{E}}=\bigcup_{e\in\mathcal{E}}\mathcal{O}_e$. In this manner, we define
a specific metric of locality of $f^{*}$ regarding $\mathcal{E}$ as follows:
\begin{equation}
            \textbf{Loc}(f^{*}, f; \mathcal{O}_{e}) = \mathbb{E}_{x \in\mathcal{O}_{e} } \mathbbm{1}\{f^*(x)= f(x)\},
\end{equation}
\begin{equation}
        \textbf{Loc}(f^{*}, f; \mathcal{O}_{\mathcal{E}})=\mathbb{E}_{x \in \mathcal{O}_{\mathcal{E}}} \mathbbm{1}\{f^*(x)= f(x)\}, \ \ \text{where}\ \ \mathcal{O}_{\mathcal{E}}=\bigcup_{e\in\mathcal{E}}\mathcal{O}_e.
\end{equation}

%\begin{itemize}
%    \item \textbf{Collective Locality.} Collective Locality (CL) is defined as the expectation of unmatched agreement between the edited model and unedited model for out-scope inputs of a series of edits $\mathcal{E}$. GL measures the method's ability to maintain locality from a global view that involves multiple edits.
%\begin{equation}
%    \textbf{CL}(\mathcal{E})=\mathbb{E}_{x \notin \mathcal{X}_{\mathcal{E}}} \mathbbm{1}\{f^*(x)\neq f(x)\} .
%\end{equation}
%    \item \textbf{Individual Locality.} Individual Locality (IL) is defined as the expectation of unmatched agreement between the edited model and unedited model for out-scope inputs of an edit $e$. This metric is used to evaluate the degree to which a specific edit will influence the out-scope inputs.
%\begin{equation}
%    \textbf{IL}(e)=\mathbb{E}_{x\notin{\mathcal{X}_e}} \mathbbm{1}\{f^*(x)\neq f(x)\}.
%\end{equation}
%\end{itemize}
%xhibit consistent and altered behavior not just for a specific input, but also for inputs that share semantic similarities. Rather than being confined to a narrow or exact input, an edited model with strong 

\subsection{Generality}
Aside from locality, another crucial metric is \textit{\textbf{generality}}, which indicates the capability of the edited model $f^{*}$ to correctly respond to semantically similar prompts~{\cite{zhu2020modifying,chen2020recall,sharma2024locating,ni2023forgetting,mitchell2022fast}}. This requires the generalization of the updated knowledge to other in-scope inputs that do not appear in the training set while conveying similar or related meanings~{\cite{gupta2024unified, wei2024mlake}}. As such, ensuring the generality of edited models prevents the edited model from overfitting to a particular input~{\cite{zhang2024comprehensive}}.
Specifically, in the scenarios of knowledge-based model editing, the inherent diversity of natural language determines that various in-scope inputs $x$ can correspond to a specific knowledge triple $t$~\cite{wang2023easyedit}. These semantically equivalent inputs can involve differences in aspects such as syntax, morphology, genre, or even language. Existing works mostly pre-define a specific in-scope input space of each edit via different strategies~{\cite{yoon2024bigger,hu2024wilke,wu2024updating,p1,li2024unveiling}}. For example, in the CounterFact dataset proposed in \textbf{ROME}~\cite{meng2022locating}, the authors utilize prompts that involve distinct yet semantically related subjects as the in-scope input. %In \textbf{MEMIT}~\cite{meng2023mass}, the authors generate rephrasings of the original statement. In \textbf{KE}~\cite{cao2021editing}, the authors apply back-translation to achieve semantically equivalent inputs. 
%Here we denote the in-scope input space of an edit $e$ as $\mathcal{I}_e$. It is noteworthy that 
In general, the generality of an edited model $f^{*}$ is defined as the expectation of exact-match agreement between the output of the edited model and true labels for in-scope inputs, which can be defined on either an edit $e$ or the edit set $\mathcal{E}$ as:
\begin{equation}
        \textbf{Gen}(f^*; e)=\mathbb{E}_{x \in \mathcal{X}_{{e}}} \mathbbm{1}\{f^*(x)\in \mathcal{Y}_e^*\},
\end{equation}
\begin{equation}
        \textbf{Gen}(f^*; \mathcal{E})=\mathbb{E}_{x \in \mathcal{X}_{\mathcal{E}}} \mathbbm{1}\{f^*(x)\in \mathcal{Y}_e^*\}, \ \ \text{where}\ \ \mathcal{X}_{\mathcal{E}}=\bigcup_{e\in\mathcal{E}}\mathcal{X}_e.
\end{equation}
%Note that, unlike locality, the measurement of generality does not possess a universal version, as the in-scope inputs are always pre-defined in each individual dataset.

\subsection{{Portability}}
{
In addition to generality, another vital metric is \textit{\textbf{portability}}, which measures the effectiveness of the edited model $f^{*}$ in transferring a conducted edit to other \textit{logically related edits} that can be interpreted via reasoning~\cite{zhang2024comprehensive}. For example, if an edit is conducted towards the President of the USA, the edit regarding the query ``Which political party does the current President of the USA belong to?'' should also be achieved. This ensures that the edited model is not limited to responding to specific input formats.  In concrete, the transfer of knowledge is crucial for robust generalization of the edited model. In practice, portability can be assessed with logically related edits obtained in different ways~\cite{yao2023editing,cohen2024evaluating}. Denoting an edit as $e=(s,r,o\rightarrow o^*)$, hereby we introduce two common types of logically related edits $\tilde{e}$. (1) Reversed Relation: $\tilde{e}=(o\rightarrow o^*, \tilde{r},s)$, where $\tilde{r}$ is the reversed relation of $r$, and (2) Neighboring Relation: $\tilde{e}=(s, r\oplus r_\epsilon, \epsilon \rightarrow \epsilon^*)$, where both $(o, r_\epsilon, \epsilon)$ and $(o^*, r_\epsilon, \epsilon^*)$ exist in the pre-trained knowledge, and $r\oplus r_\epsilon$ is a combined relation from $r$ and $r_\epsilon$. In this manner, we define portability as the edited model performance on one or multiple logically related edits as follows:
%(1) \textit{Subject Replace}: This tests whether the model can generalize the edited attribute to other descriptions of the same subject by replacing the subject in the question with an alias or synonym~{\cite{DeCao2021}}. (2) \textit{Reversed Relation}: When the target of a subject and relation is edited, this aspect tests the model's ability to handle the reversed question to ensure the target entity's attributes are also updated~{\cite{DeCao2021}}. (3) \textit{One-hop}: This evaluates whether the modified knowledge is usable by the model for downstream tasks by constructing a reasoning dataset that tests the model's ability to apply the edited knowledge to related questions~{\cite{Thibodeau2022}}. Ensuring the portability of edited models is crucial for verifying the implications of an edit for realistic applications and robust generalization~{\cite{Thibodeau2022}}.
\begin{equation}
        \textbf{Por}(f^*; \tilde{e})=\mathbb{E}_{x \in \mathcal{X}_{\tilde{e}}} \mathbbm{1}\{f^*(x)\in \mathcal{Y}_{\tilde{e}}^*\},
\end{equation}
\begin{equation}
        \textbf{Por}(f^*; \widetilde{\mathcal{E}})=\mathbb{E}_{x \in \mathcal{X}_{\widetilde{\mathcal{E}}}} \mathbbm{1}\{f^*(x)\in \mathcal{Y}_{\tilde{e}}^*\}, \ \ \text{where}\ \ \mathcal{X}_{\widetilde{\mathcal{E}}}=\bigcup_{\tilde{e}\in\mathcal{\widetilde{E}}}\mathcal{X}_{\tilde{e}}.
\end{equation}
}

\subsection{Retainability}

\textbf{\textit{Retainability}} characterizes the ability of KME techniques to preserve the desired properties of edited models after \textbf{multiple consecutive edits}~{\cite{jiang2024learning,yu2024melo,gu2023pokemqa}}. In the presence of ever-evolving information, practitioners may need to frequently update a conversational model (i.e., sequential editing). Such a KME setting requires that the model does not forget previous edits after each new modification~\cite{e1}. {It is essential to distinguish retainability from scalability}, which evaluates the model's ability to handle a vast number of edits~\cite{p4}. In contrast, retainability assesses the consistent performance of the model after each individual edit, presenting a more challenging objective to achieve. Recently, \textbf{T-Patcher}~\cite{Transformer-Patcher} first explores the sequential setting of KME and observes that many existing approaches significantly fall short in terms of retainability. In \textbf{SLAG}~\cite{hase2023methods}, the authors also discover a significant drop in editing performance when multiple beliefs are updated continuously. To assess the retainability of an edited language model $f^{*}$, we define it as follows:
%Retainability refers to the property of language model editing techniques that can desirable properties of edited models after a series of edits. For example, due to the constantly emerging information, the practitioners need to continuously update (i.e., sequential editing) for a conversation model, requiring that after each edit the previous edits are not forgotten. Therefore, retainability remains a crucial property for the practicality of KME techniques. Not that this is different from the scalability, which is intended to measure the capability of enduring a large number of edits. Distinctly, retainability measures the overall performance after each edit, which is more difficult to achieve. Recently, T-Patcher~\cite{Transformer-Patcher} first explores the sequential setting of KME and observes that existing works signinfcatly performs badly in retainability. To effectively evaluate retainability of edited language models, we define the retainabltliy of an edited model as follows:
\begin{equation}
\begin{aligned}
    \mathbf{Ret}(M;\mathcal{E})=\frac{1}{|\mathcal{E}|-1}\sum\limits_{i=1}^{|\mathcal{E}|-1}\mathbf{Acc}(M(f;\{e_1,e_2,\dotsc, e_{i+1}\})) - \mathbf{Acc}(M(f;\{e_1,e_2,\dotsc, e_{i}\}))    
\end{aligned}
\end{equation}
where $\mathbf{Acc}$ is the accuracy measurement, $|\mathcal{E}|$ is the number of edits in the edit set, and $M$ denotes the editing strategy that modifies the pre-trained model $f$ into $f^{*}$ with $i/i+1$ consecutive edits $\{e_1,e_2,\dotsc, e_{i}, (e_{i+1})\}$. The retainability metric aims to quantify the effect of applying consecutive edits to a model and measures how the performance will change the editing strategy $M$, where a higher retainability means that after each edit, the less the change in the overall performance of the edited model $f^{*}$ is required. 

\subsection{Scalability}
The \textbf{\textit{scalability}} of an editing strategy refers to its capability to incorporate a large number of edits simultaneously~\cite{p4}.
Recently, several works have emerged that can inject multiple new knowledge into specific parameters of pre-trained LLMs~\cite{yoon2024bigger,zhang2024comprehensive}. For instance, {\textbf{SERAC}}~\cite{SERAC} can perform a maximum of 75 edits. In addition, \textbf{MEMIT}~\cite{meng2023mass} is proposed to enable thousands of edits without significant influence on editing accuracy. When there is a need to edit a model with a vast number of edits concurrently, {simply employing the current knowledge-based model editing techniques in a sequential manner is proven ineffective in achieving such scalability~\cite{yao2023editing}}. To effectively evaluate the scalability of edited language models, we define the scalability of an edited model as follows:
\begin{equation}
    \mathbf{Sca}(M;\mathcal{E})=\mathbb{E}_{e\in\mathcal{E}}\mathbf{Acc}(M(f;e)) -\mathbf{Acc}(M(f;\mathcal{E})),
    \label{eq:delta_scala}
\end{equation}
where $\mathbf{Acc}(M(f;\mathcal{E}))$ denotes the accuracy of the edited model after conducting all edits in $\mathcal{E}$, whereas $\mathbf{Acc}(M(f;e))$ is the accuracy of only performing the edit $e$. $\mathbf{Sca}$ demonstrates the model performance and practicality in the presence of multiple edits. Nevertheless, we note that baseline value $\mathbf{Acc}(M(f;\{e\}))$ is also important in evaluating the scalability of various models. This is because, with higher accuracy for each $e$, the retainment of such performance after multiple edits is more difficult. Therefore, we further define the relative version of Eq.~(\ref{eq:delta_scala}) as follows:
\begin{equation}
        \mathbf{Sca}_{rel}(M;\mathcal{E})=\left(\mathbb{E}_{e\in\mathcal{E}}\mathbf{Acc}(M(f;\{e\})) -\mathbf{Acc}(M(f;\mathcal{E}))\right)/\mathbb{E}_{e\in\mathcal{E}}\mathbf{Acc}(M(f;\{e\})).
\end{equation}
The introduced scalability measurement further considers the magnitude of the original accuracy to provide a fairer evaluation.
\input{category}

\section{Methodologies}\label{sec:method}

In this section, we introduce existing knowledge-based model editing (KME) strategies in detail. We first provide an innovative taxonomy of existing KME strategies based on how and where the new knowledge is injected into the pre-trained LLMs, where the advantages and drawbacks are thoroughly discussed. We then introduce various methods from each category, with an emphasis on analyzing the technical details, insights, shortcomings, and their relationships.

\subsection{Categorization of KME Methods}
\label{sec:category}

\begin{figure}[!t]
  \centering
  %\vspace{-0.15in}
  \includegraphics[width=0.98\textwidth]{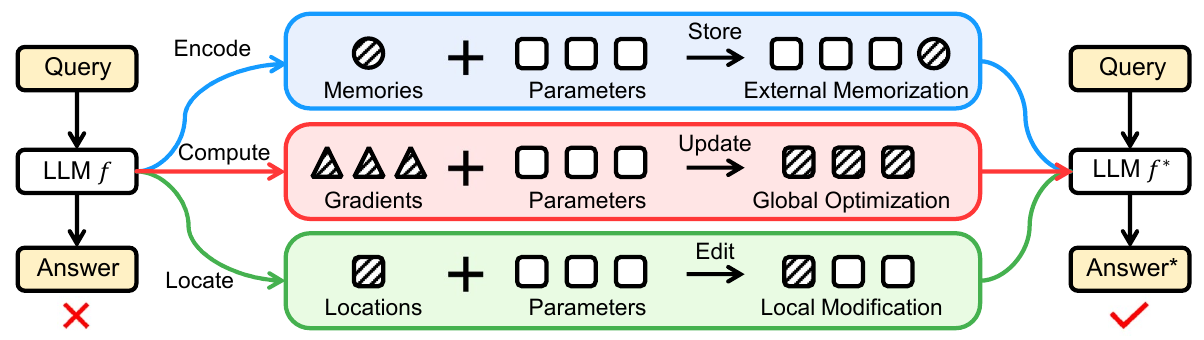}
  \caption{The illustration of three categories of KME methods: \textbf{External Memorization}, \textbf{Global Optimization}, and \textbf{Local Modification}.}
  \vspace{-0.15in}
        \label{fig:illustration}
\end{figure}

Faced with the rapid deprecation of old information and the emergence of new knowledge,  various KME methodologies have been proposed to update the pre-trained LLMs to maintain their updatedness and relevancy. KME ensures that new knowledge can be efficiently incorporated into the pre-trained LLMs without negatively influencing the pre-trained knowledge irrelevant to the edit. In this survey, we categorize existing KME methods into three main classes as follows:

\begin{itemize}[leftmargin=0.35cm]
    \item \textbf{External Memorization}-based methods \textbf{leverage an external memory} to store the new knowledge for editing without modifying the pre-trained weights, where the pre-trained knowledge can be fully preserved in the LLM weights. By storing new knowledge with external parameters, the memory-based strategies enable precise representation of new knowledge with good scalability, as the memory is easily extensible to incorporate new knowledge.

    \item \textbf{Global Optimization}-based methods seek to \textbf{achieve generalizable incorporation} of the new knowledge into pre-trained LLMs via optimization with the guidance of new knowledge, {where tailored strategies are introduced to limit the influence of other pre-trained knowledge}, distinguishing it from naive fine-tuning. Nevertheless, these methods may fall short in editing efficiency when applied to LLMs due to the large number of parameters to be optimized.

    \item \textbf{Local Modification}-based methods aim to \textbf{locate the related parameters} of specific knowledge in LLMs and update it accordingly to incorporate the new knowledge relevant to the edit. The main advantage of local modification is the possibility of only updating a small fraction of model parameters, {thereby providing considerable memory efficiency compared to memorization-based methods and computational efficiency compared to global optimization.} 
    %These methods are also naturally suitable for causal analysis, as they can identify the most responsible parameters for specific knowledge.
    % Based on gradients calculated from the desirable knowledge, these methods optimize the located neurons to perform editing.  
\end{itemize}
The above categorization is achieved based on \textbf{where} (e.g., external parameters or internal weights) and \textbf{how} (e.g., via optimization or direct incorporation) new knowledge is introduced into the LLM during editing. Methods in each category exhibit different strengths and weaknesses regarding the four crucial evaluation metrics introduced in Sec.~\ref{sec:metric}. For example, external memorization prevails in scenarios that require massive editing while the computational resources are limited, as the size of the memory is controllable to fit into different requirements. On the other hand, global optimization is advantageous when practitioners focus more on the generality of edited knowledge, as the optimization can promote the learning of relevant knowledge~\cite{aghajanyan-etal-2021-intrinsic}. The taxonomy is visually illustrated in Fig.~\ref{fig:categorization}, and a more detailed demonstration of each category is presented in Fig.~\ref{fig:illustration}. 
%The specific characteristics of all methods are summarized in Table~\ref{tab:comparison}. 

%In this survey, we introduce and compare the pros and cons of different KME methods from these three perspectives. It is also noteworthy that these perspectives are not mutually exclusive, as the combination of any two is possible. Before introducing more details of each category, we provide further details of methods in each categorization in Fig.~\ref{fig:illustration}.

%    \item External Memorization (Scalability, Specification)
%    \item Global Optimization (?) (Scalability, Retention)
%    \item Local Modification (Specification, Retention)

%%Yaochen: Removed due to repetition
%With proper techniques for encoding knowledge in parameters, this type of methods can store a considerable amount of edits with a simple way for mapping knowledge into outputs,

%\begin{equation}
%        \begin{aligned}
%            & \min_\omega \mathbb{E}_{(x,y \rightarrow y^{*}) \sim P_\mathcal{E}} \mathcal{L} (f^*_{\phi, \omega}(x), y^*),where\ \  f^*_{\phi, \omega}=M(f_\phi, \omega; \mathcal{E}),\\
%        &s.t. f^*_{\phi, \omega}(x)=y,\ \  \forall (x, y)\sim P_\mathcal{F},
%        %+ \mathbb{E}_{(x, y)\sim P_\mathcal{F}}R(f(x), y),
%    \end{aligned}
%\end{equation}

\subsection{External Memorization}\label{sec:external}
\subsubsection{\textbf{Overview}} The editing approaches via external memorization aim to modify the current model $f_\phi$ (with parameter $\phi$) via introducing external memory represented by additional trainable parameters $\omega$ that encodes the new knowledge, resulting in an edited LLM model $f^*_{\phi, \omega}$. The rationale behind the external memorization strategy is that storing new knowledge in additional parameters is intuitive and straightforward to edit the pre-trained LLMs with good scalability, as the parameter size can be expanded to store more knowledge. In addition, the influence on the pre-trained knowledge can be minimized as this strategy does not alter the original parameters $\phi$. Based on the general formulation of KME in Eq. (\ref{def:general}), the objective of external memorization approaches can be formulated as follows:
    \begin{equation} \label{eq:formulation}
    \begin{aligned}
            & \min \mathbb{E}_{e \in \mathcal{E}} \mathbb{E}_{x, y^{*} \in \mathcal{X}_e, \mathcal{Y}^*_e} \mathcal{L} (f^*_{\phi, \omega}(x), y^{*}), \text{where}\ \  f^*_{\phi, \omega}=M(f_\phi, \omega; \mathcal{E}),\\
        &\st f^*_{\phi, \omega}(x)=f_\phi(x),\ \  \forall x\in \mathcal{X}\setminus \mathcal{X}_\mathcal{E},
        %+ \mathbb{E}_{(x, y)\sim P_\mathcal{F}}R(f(x), y).
    \end{aligned}
    \end{equation}
 where $f_\phi$ denotes the LLM before editing with the pre-trained  parameter $\phi$, and $f^*_{\phi, \omega}$ denotes the edited LLM with $\phi$ and additional parameter $\omega$ as the external memorization. Moreover, based on whether the introduced parameters are directly incorporated into the model process or not, external memorization strategies can be divided into two categories, i.e., \textbf{memory-based methods} and \textbf{extension-based methods}.

% Yaochen:
% involve the strategic removed
%Generally, Fundamental Memory-based Strategy, e.g.,

\subsubsection{\textbf{Memory-based Strategies}}
In memory-based strategies, the external memory, outside the intrinsic architecture of the pre-trained LLM, functions as a repository to store edited knowledge. {Here the edits are generally converted to text via pre-defined templates~\cite{MQuAKE, IKE,p2}.} The LLM can access and update this memory as required during inference. 

One exemplar work is \textbf{SERAC} ~\cite{SERAC}, which stores the edited samples $x, y^{*} \in \mathcal{X}_{e}, \mathcal{Y}^{*}_{e}$ in a cache without performing modifications on the original model. When presented with a new prompt $x^{\prime}$, SERAC uses a scope classifier to determine whether the prompt falls within the scope of any cached instances. If yes, the desirable output $y^{\prime}$ associated with the new prompt $x^{\prime}$ is predicted via a counterfactual model $f_c$ which utilizes the most relevant edit example as follows:
\begin{equation}
      f^*_{\phi,\omega}(x)
      =\left\{\begin{array}{ll}
      f_{\phi}(x),  & \text{if}\  x\ \text{is not in scope of any edit},\\
      f_c(x,\mathcal{E}), & \text{otherwise}.\\
      \end{array}\right.
\end{equation}
SERAC is a gradient-free approach to KME without relying on gradients of the target label $y^{*}$ w.r.t. the pre-trained model parameters. In addition to using memory as an external repository, the desirable edits can also be stored in the form of human feedback. For example, ~\textbf{Language Patch}~\cite{Patch} performs editing by integrating patches in natural language, and \textbf{MemPrompt}~\cite{MemPrompt} involves human feedback prompts to address the issue of lacking commonsense knowledge regarding a particular task. An integral feature of the {Language Patch}~\cite{Patch} framework is its ability to empower practitioners with the capability to create, edit, or remove patches without necessitating frequent model re-training. This trait not only streamlines the development process but also enhances the adaptability and versatility of the edited model. To enable the automatic correction in memory, {MemPrompt}~\cite{MemPrompt} equips the language model with a memory bank containing corrective feedback to rectify misunderstandings. Specifically, MemPrompt leverages question-specific historical feedback to refine responses on novel and unencountered instances through prompt adjustments.

In \textbf{KAFT}~\cite{KAFT}, controllability is achieved through the utilization of counterfactual data augmentations. In this approach, the entity representing the answer within the context is substituted with an alternative but still plausible entity. This substitution is intentionally designed to introduce a conflict with the genuine ground truth, thereby enhancing the controllability and robustness of LLMs with respect to their working memory. The aim is to ensure that LLMs remain responsive to pertinent contextual information while filtering out noisy or irrelevant data.

%Another work \textbf{KAFT}~\cite{KAFT} ensures controllability by employing counterfactual data augmentations, where the answer entity within the context is replaced with an alternative yet plausible entity. The entity is intended to conflict with the actual ground truth, thus enhancing the controllability and robustness of LLM working memory by ensuring its responsiveness to relevant context while disregarding noise or irrelevant information. 

In addition to relying on parameter-based memory, recent works also leverage prompting techniques of LLMs, e.g., in-context learning~\cite{dong2022survey} and chain-of-thought prompting~\cite{wei2022chain}, to promote editing performance of external memorization.
Specifically,
\textbf{IKE}~\cite{IKE} introduces novel factual information into a pre-trained LLM via in-context learning, where a set of $k$ demonstrations, i.e., $\omega=\{x_{i}, y^{*}_{i}\}_{i=1}^{k}$, is selected as the reference points. These demonstrations will alter the prediction of a target factual detail when the input is influenced by an edit. Particularly, IKE guarantees a balance between generality and locality via storing factual knowledge as prompts. The process can be formulated as follows:
\begin{equation}
f^*_{\phi, \omega}(x)=f_\phi( \omega \| x),\ \text{where}\ \omega=\{x_{i}, y^{*}_{i}\}_{i=1}^{k}.
\end{equation}
Here $\|$ denotes the concatenation of the reference points in $\omega$ and the input $x$, which follows an in-context learning manner. Note that in this process, the framework first transforms all new facts into natural language to input them into LLMs. {Similar methods of knowledge editing based on prompts~\cite{p1,p2,p3,p4} can also update and modify knowledge within large language models (LLMs). These approaches allow users to guide the model to generate desired outputs by providing specific prompts, and effectively and dynamically adjusting the model's knowledge base. By leveraging the flexibility of prompts and the contextual understanding of LLMs, users can correct or update information in real-time. These methods offer immediacy, flexibility, and cost-efficiency, making them powerful tools for maintaining the accuracy and relevance of language models in rapidly evolving knowledge domains.} Although the prompt approaches effectively edit factual knowledge via in-context learning, they cannot solve more complex questions that involve multiple relations. To deal with this, \textbf{MeLLo}~\cite{MQuAKE} first explores the evaluation of the editing effectiveness in language models regarding multi-hop knowledge. For example, when editing knowledge about the president of the USA, the query regarding the president's children should change accordingly. %Specifically, MQuAKE aims to assess whether edited models can correctly answer questions that represent such chains of facts. 
MeLLo proposes to enable multi-hop editing by breaking down each query into subquestions, such that the model generates a provisional answer. Subsequently, each subquestion is used to retrieve the most pertinent fact from the memory to assist the model in answering the query. 
%The language model then assesses whether the retrieved fact contradicts the provisional answer and adjusts the prediction accordingly. 

\subsubsection{\textbf{Extension-based Strategies}}
Extension-based strategies utilize supplementary parameters to assimilate modified or additional information into the original language model. These supplementary parameters are designed to represent the newly introduced knowledge or necessary adjustments tailored for specific tasks or domains. Different from memory-based methods, by incorporating new parameters into the language model, extension-based approaches can effectively leverage and expand the model's functionality. %, enabling it to seamlessly integrate the updated or domain-specific information.

Extension-based methods can be implemented through various means, and one representative way is to modify the Feed-forward Neural Network (FFN) output. For example, \textbf{CALINET}~\cite{CALINET} uses the output from sub-models fine-tuned specifically on factual texts to refine the original FFN output produced by the base model. %resulting in necessary adjustments. 
Another technique \textbf{T-Patcher}~\cite{Transformer-Patcher} introduces a limited number of trainable neurons, referred to as ``patches,'' in the final FFN layer to alter the model's behavior while retaining all original parameters to avoid reducing the model's overall performance. Generally, these methods that refine the structure of FFN can be formulated as follows:
\begin{equation}
\operatorname{FFN}(\textbf{h}) =\operatorname{GELU}\left(\textbf{h} \mathbf{W}_1\right) \mathbf{W}_2+ \operatorname{GELU}\left(\bh \cdot \mathbf{k}_p +b_p\right)\cdot \mathbf{v}_p,
\end{equation}
where $\mathbf{k}_p$ is the patch key, $\mathbf{v}_p$ is the patch value, and $b_p$ is the patch bias scalar. The introduced patches are flexible in size and can be accurately activated to edit specific knowledge without affecting other model parameters.

%Alternatively, an adapter can be incorporated into a selected layer of a pre-trained model, which contains a discrete dictionary of keys and values. Here each key is a cached activation predicted by the previous layer, and each value is a corresponding output that decodes into the desired model output. The dictionary is regularly updated over time. With this thought, an approach \textbf{GRACE}~\cite{hartvigsen2022aging} adds an adapter to make informed decisions about whether to utilize the dictionary for a given input by employing a deferral mechanism. Balancing the benefits of preserving the original model's integrity with the practical considerations of storage space is a key consideration when employing this approach.

Alternatively, a different technique involves integrating an adapter into a specific layer of a pre-trained model. This adapter consists of a discrete dictionary comprising keys and values, where each key represents a cached activation generated by the preceding layer and each corresponding value decodes into the desired model output. This dictionary is systematically updated over time. In line with this concept, \textbf{GRACE}~\cite{hartvigsen2022aging} introduces an adapter that enables judicious decisions regarding the utilization of the dictionary for a given input, accomplished via the implementation of a deferral mechanism. It is crucial to achieve a balance between the advantages of preserving the original model's integrity and the practical considerations associated with storage space when implementing this approach. {\textbf{COMEBA-HK}~\cite{e1} incorporates hook layers within the neural network architecture. These layers allow for the sequential editing of the model by enabling updates to be applied in batches. This approach facilitates the integration of new knowledge without requiring extensive retraining of the entire model, making it a scalable solution for continuous learning and adaptation. \textbf{SWEA}~\cite{e2} focuses on altering the embeddings of specific subject words within the model. By directly updating these embeddings, the method can inject new factual knowledge into the LLMs. This approach ensures that the updates are precise and relevant, thereby enhancing the model's ability to reflect current information accurately.}

\subsubsection{\textbf{Summary}} The eternal memorization methodology operates by preserving the parameters within the original model while modifying specific output results through external interventions via memory or additional model parameters. One notable advantage of this approach is its minimal perturbation of the original model, thereby ensuring the consistency of unedited knowledge. It allows for precise adjustments without necessitating a complete overhaul of the model's architecture. However, it is imperative to acknowledge a trade-off inherent in this methodology. Its efficacy is contingent upon the storage and invocation of the edited knowledge, a factor that leads to concerns regarding storage capacity. Depending on the scale of knowledge to be edited, this approach may entail substantial storage requisites. Therefore, cautiously seeking a balance between the advantages of preserving the original model's integrity and the practical considerations of storage capacity becomes a pivotal concern when employing this particular approach.

%This method relies on storing and invoking the edited knowledge, which can lead to concerns related to storage space. Depending on the volume, this approach might result in significant storage requirements. Balancing the benefits of preserving the original model's integrity with the practical considerations of storage space is a key consideration when employing this approach.

\subsection{Global Optimization}\label{sec:global}
\subsubsection{{\textbf{Overview}}}
% TODO: 为什么不用fine-tuning
% YC: Global optimization = finetuning + locality constraints
Different from external memorization methods that introduce new parameters to assist the editing of pre-trained LLMs, there also exist branches of works that do not rely on external parameters or memory. Concretely, global optimization strategies aim to inject new knowledge into LLMs by updating all parameters, i.e., $\phi$ in Eq. (\ref{eq:formulation}). Through fine-tuning model parameters with {specific designs to ensure the preservation of knowledge irrelevant to the target knowledge $t^*$}, the LLMs are endowed with the ability to absorb new information without altering unedited knowledge. Generally, the goal of global optimization methods can be formulated as follows:
    \begin{equation}
    \begin{aligned}
            & \min \mathbb{E}_{e \in \mathcal{E}} \mathbb{E}_{x, y^{*} \in \mathcal{X}_e, \mathcal{Y}^*_e} \mathcal{L} (f_{\phi^*}(x), y^{*}),\ \text{where}\ \  f_{\phi^*}=M(f_\phi; \mathcal{E}),\\
        &\st f_{\phi^*}(x)=f_{\phi}(x),\ \  \forall x\in \mathcal{X}\setminus \mathcal{X}_\mathcal{E},
        %+ \mathbb{E}_{(x, y)\sim P_\mathcal{F}}R(f(x), y),
    \end{aligned}
    \label{eq:finetune}
    \end{equation}
 where $f_\phi$ denotes the LLM before editing with the pre-trained parameter $\phi$, and $f_{\phi^*}$ denotes the edited LLM with updated parameter $\phi^*$.
Generally, these methods focus more on the precision and generality of desirable knowledge, as the fine-tuning process ensures that the LLMs achieve satisfactory results regarding the edits and relevant knowledge. Nevertheless, as fine-tuning affects all parameters, they cannot easily preserve the locality of edited models, i.e., maintaining consistent output for unedited knowledge~\cite{yao2023editing}. In practice, directly applying fine-tuning strategies typically exhibits suboptimal performance on KME due to overfitting concerns~\cite{wang2023easyedit,meng2023mass}. Furthermore, fine-tuning large language models is also time-consuming and lacks scalability for multiple edits. Therefore, recently, motivated by these two challenges in fine-tuning, several global optimization works have been proposed and can be categorized as \textbf{constrained fine-tuning methods} and \textbf{intermediate fine-tuning methods}.
{Note that this section primarily focuses on methods from the model training perspective. Additionally, certain studies~\cite{jiang2024learning, gangadhar2024modeleditingstandardfinetuning} address the overfitting challenge by constructing more a comprehensive $\mathcal{X_{E}'}$ with the following fine-tuning goal:}
\begin{equation}
    \begin{aligned}
            & \min \mathbb{E}_{e \in \mathcal{E}} \mathbb{E}_{x, y^{*} \in \mathcal{X}_{e}', {\mathcal{Y}^*_e}' } \mathcal{L} (f_{\phi^*}(x), y^{*}),\ \text{where}\ \  f_{\phi^*}=M(f_\phi; \mathcal{E}),\\
        &\st \mathcal{X_E}\subset \mathcal{X_{E}}',  \mathcal{X_{E}}'\subseteq\mathcal{X}.
        %+ \mathbb{E}_{(x, y)\sim P_\mathcal{F}}R(f(x), y),
    \end{aligned}
    \label{eq:finetune}
    \end{equation}

%into two claessit is imperative to devise specific constraints for global optimization to tackle these challenges in KME.

% As a common and effective technique in Natural Language Processing (NLP) tasks, fine-tuning (FT) is used to transfer knowledge from the pre-trained model to custom domains. FT involves taking a pre-trained model that has learned general patterns and features from a large dataset and then further training it on a smaller, task-specific dataset. 

%The scenario of language model editing is quite similar to transfer learning, in which updated knowledge plays the role of a task-specific dataset. As a result, those existing methods that apply pre-trained large language models (LLM) to specific domains through fine-tuning are, in principle, applicable to model editing. However, in the context of model editing, it is desirable to avoid impacting the performance of the model on the unmodified knowledge. 
%We find that conventional fine-tuning, due to overfitting concerns, exhibits suboptimal performance on these tasks. Furthermore, fine-tuning large language models is time-consuming and lacks scalability.
\subsubsection{\textbf{Constrained Fine-tuning}} Constrained fine-tuning strategies generally apply specific constraints to prevent updating on non-target knowledge in $\{\mathcal{X}\setminus \mathcal{X}_\mathcal{E},\mathcal{Y}\setminus \mathcal{Y}_\mathcal{E}\}$. In this manner, the objective in Eq.~(\ref{eq:finetune}) is transformed into a constrained optimization problem:
\begin{equation}
    \begin{aligned}
    & \min \mathbb{E}_{e \in \mathcal{E}} \mathbb{E}_{x, y^{*} \in \mathcal{X}_e, \mathcal{Y}^*_e} \mathcal{L} (f_{\phi^*}(x), y^{*}),\ \text{where}\ \  f_{\phi^*}=M(f_\phi; \mathcal{E}),\\
    & \st\ \|\mathcal{L}(f_{\phi^*}(x), y)-\mathcal{L}(f_{\phi}(x), y)\| \leq \delta, \forall x,y\in \mathcal{X}\setminus \mathcal{X}_\mathcal{E},\mathcal{Y}\setminus \mathcal{Y}_\mathcal{E},
    \end{aligned}
        \label{eq:dft}
\end{equation}
where $\phi$, $\phi^*$ are\ the\ parameters\ before\ and\ after\ updating, respectively. $\delta$ is a scalar hyper-parameter to restrict the difference between losses of $f_{\phi^*}$ and $f_\phi$. The constraint in Eq.~(\ref{eq:dft}) restricts the change of the edited model on unmodified knowledge. Zhu et al. \cite{zhu2020modifying} first propose an approximate optimization constraint that is easier for implementation and computation:
\begin{equation}
    \begin{aligned}
    & \min \mathbb{E}_{e \in \mathcal{E}} \mathbb{E}_{x, y^{*} \in \mathcal{X}_e, \mathcal{Y}^*_e} \mathcal{L} (f_{\phi^*}(x), y^{*}),\ \text{where}\ \  f_{\phi^*}=M(f_\phi; \mathcal{E}),\\
    & \st\ \|\phi^*-\phi\| \leq \delta.
    \end{aligned}
  \label{eq:16}
\end{equation}
The updates are regularized by restricting the norm of parameters before and after updating. 
{
\textbf{RECT}~\cite{gu2024model} adopts a similar yet simpler approach, specifically modifying only the top-k\% of parameters with the largest numerical updates during fine-tuning.}
Although restricting the norm is helpful in preventing the forgetting of original knowledge, the fine-tuning process can be less effective. To deal with this, \textbf{RecAdam}~\cite{chen2020recall}, in addition to the norm constraint, applies an annealing technique to control the ratio between the parameter norm and the fine-tuning loss as follows:
\begin{equation}
    \mathcal{L}_{total}=\lambda(t)\mathcal{L}_{FT}+(1-\lambda(t))\|\phi^*-\phi\|,\ \ \text{where}\ \ \lambda(t)=\frac{1}{1+\exp(-k\cdot(t-t_0))}.
\end{equation}
Here $k$ and $t_0$ are hyper-parameters. $t$ is the number of fine-tuning steps. Such a design enables a gradual fine-tuning process that prevents massive parameter updates at the beginning. Motivated by the intuition of regularization to preserve original knowledge, \textbf{PPA}~\cite{lee2022plugandplay} employs LoRA~\cite{hu2021lora} in the feed-forward (FFN) layers of the transformer decoder. LoRA is proposed to train the expansion/reduction matrix, instead of the model parameter $\phi$, to improve training speed by only updating parameters with a low intrinsic rank via dimensionality reduction. PPA leverages plug-in modules trained with constraints via LoRA to keep original knowledge intact. 
Moreover, the authors assess whether the content of the inputs falls within the scope of $\mathcal{X}_\mathcal{E}$ using the K-adapter module~\cite{wang2020k}, and redirect such inputs to the new plug-in modules. This information is then used to determine whether to employ LoRA within the FFN layers.
{Furthermore, \textbf{MELO}~\cite{yu2024melo} clusters the edits and employs multiple non-overlapping LoRA blocks for fine-tuning each cluster separately, thereby mitigating the issue of catastrophic forgetting.}
\textbf{F-Learning} (Forgetting before Learning)~\cite{ni2023forgetting} proposes another approach to preserve original knowledge, which learns knowledge parameters $\Delta\phi$ that indicates old knowledge to be forgotten, defined as follows:
\begin{equation}
   \phi^*=\phi-\lambda\Delta\phi,\ \ \text{where}\ \   \Delta\phi=\text{FT}(\phi; \mathcal{K}_{old})-\phi.
\end{equation}
Here $\mathcal{K}_{old}$ denotes the dataset composed of old knowledge that we desire to forget, and $\text{FT}(\phi;\mathcal{K}_{old})$ is the supervised fine-tuning process of parameters $\phi$ on dataset $\mathcal{K}_{old}$. $\lambda$ is a hyper-parameter used to control the rate of forgetting. Based on the assumption that subtracting the parameters $\Delta\phi$ from $\phi$ can help the model forget this part of old knowledge~\cite{ilharco2023editing}, F-Learning defines the forgetting process as a subtraction operation to obtain the updated model parameter $\phi^*$.

%To further improve training speed, \textbf{LoRA} \cite{hu2021lora} proposes to train the expansion/reduction matrix instead of the model parameter $\phi$. The authors assume that the update of parameters possesses a low intrinsic rank. To simulate the intrinsic rank, they introduce a bypass mechanism alongside the original language model to perform a dimensionality reduction followed by dimensionality expansion. Specifically, they reformulate $\mathbf{P}\phi^{'(d_s)}$ in Eq.~(\ref{eq:subspace}) into two components: matrix $\mathbf{B}$ (dimensionality expansion) and $\mathbf{A}$ (dimensionality reduction). Thus, they have:
%\begin{equation}
%    \phi^*x=(\phi + \mathbf{B}\mathbf{A})x, x\in \mathcal{X}.
%\end{equation}
%Note that the language model parameter $\phi$ remains unchanged during the fine-tuning process. Due to its separable nature, LoRA can be easily applied to various language models. 

%Despite direct fine-tuning, more recent works~\cite{aghajanyan-etal-2021-intrinsic,qin2022exploring} also introduce \textbf{meta-learning} methods and \textbf{subspace fine-tuning methods} to address the issue of overfitting and improve scalability.

On the other hand, other works also resort to meta-learning~\cite{finn2017model,vanschoren2018meta} to apply more flexible constraints. 
Meta-learning addresses the issue of overfitting by training a model that can quickly adapt to new tasks~\cite{hospedales2021meta}. By exposing the model to a variety of tasks during training, meta-learning improves the model's ability to generalize from limited data and reduces the risk of overfitting individual tasks~\cite{huisman2021survey}. In the scenario of KME, the optimal model parameters $\phi^*$ should minimize the expected loss over a variety of meta-tasks \cite{ravi2016optimization}:
\begin{equation}
    \phi^* = \argmin_\phi\mathbb{E}_{D\sim\mathcal{D}}[\mathcal{L}_\phi({D})],
\end{equation}
where $\mathcal{D}$ corresponds to the sample set for each meta-task $D$. Moreover, each meta task ${D}$ contains multiple $(x^*, y^*)$ pairs for editing. In practice, such methods often introduce additional objective functions or networks to regulate parameter updates. As a typical meta-learning method for KME, \textbf{Editable Training}~\cite{sinitsin2020editable} focuses on effectively rectifying errors within models while preserving their performance on other irrelevant data instances. Following a model-agnostic training manner, the authors introduce additional constraints to restrict parameter updates in a different way. Specifically, the loss function is separated into $\mathcal{L}_{base}$ (task-specific objective function), $\mathcal{L}_{edit}$ (computed on the edit set $\mathcal{X}_\mathcal{E}$), and $\mathcal{L}_{local}$ (computed on samples in $\mathcal{X}\setminus \mathcal{X}_\mathcal{E}$). Moreover, the models are updated in a meta-learning manner, where $k$ steps of gradient descent would be applied for parameters before computing the objective function. 

\subsubsection{\textbf{Intermediate Fine-tuning Strategies}}
While constrained fine-tuning techniques have demonstrated remarkable efficacy in a variety of NLP tasks~\cite{wortsman2022robust,ziegler2019fine,bakker2022fine}, they still exhibit instability and high computational cost when applied to KME, primarily due to the necessity of altering all parameters~\cite{yao2023editing}. A potential solution to address this challenge is to utilize an intermediate model to obtain the updated parameters in an efficient manner. Such an intermediate model is required to maintain significantly fewer parameters to ensure efficiency~\cite{cheng2023editing}. In general, recent works have widely adopted the \textbf{Hyper-Network}~\cite{ha2016hypernetworks} as the intermediate model. Specifically, the Hyper-Network is a small network that generates the weights for a larger network, referred to as the main network. Specifically, the Hyper-Network takes inputs that contain information about the structure of the weights and generates the weights for layers in the main network. 
With the generated weights, the main network is updated to map input data to desired output targets. The updating process for the main network, denoted as $\phi$, can be defined as follows:
%While traditional meta-learning methods have achieved excellent results in the task of machine translation~\cite{gu2018meta}, they appear less stable in the case of language model editing due to the requirement for fine-tuning. \textbf{Hyper-Network}~\cite{ha2016hypernetworks} could be a solution to address this problem. The hyper-network model is a small network that generates the weights for a larger network, called a main network. The main network behaves like a usual neural network, learning to map raw inputs to desired targets. In contrast, the Hyper-Network takes inputs that contain information about the structure of the weights and generates the weights for layers in the main network. We define the updating process on the main network $\phi$ as follows:
\begin{equation}
\begin{aligned}
     \phi^*&=\phi + \Delta\phi, \ \ \text{where}\ \ 
     \Delta\phi= \text{H}(\nabla_\phi\mathcal{L} (f_{\phi}(x), y^{*})) \ \  \text{and} \ \ 
    x, y^* \in \mathcal{X}_\mathcal{E}, \mathcal{Y}^*_\mathcal{E},
\end{aligned}
\end{equation}
where $\text{H}(\cdot)$ denotes the hyper-network. $\Delta\phi$ is the weight deviation calculated by the hyper-network. According to a recent study~\cite{vonoswald2022continual}, task-specific Hyper-Networks (i.e., networks that generate target model weights based on task attributes) are effective in mitigating catastrophic forgetting issues. Therefore, such methods are suitable for the setting of KME, which requires the preservation of unedited knowledge. 

Recently, researchers have proposed to adopt hyper-networks in various ways for parameter updates in KME. As a classic example, \textbf{KE}~\cite{cao2021editing} first proposes to edit knowledge and rectify erroneous or unexpected predictions without expensive fine-tuning. Specifically, it trains a hyper-network via constrained optimization to modify facts without affecting pre-trained knowledge irrelevant to the edit. The trained hypernetwork is then used to predict the weight update at the inference time. Based on KE, \textbf{SLAG}~\cite{hase2023methods} further appends metrics for two types of input texts: ({1)} Inputs that are not in the desired edit set $\mathcal{X}_\mathcal{E}$ but logically related to $\mathcal{E}$; (2) Inputs that share a formal resemblance to edited knowledge, but do not lead to changes in the prediction outcomes. 

However, hyper-networks are generally not capable of updating large language models due to the massive parameter size. To tackle this challenge, \textbf{MEND}~\cite{mitchell2022fast} adopts a mechanism referred to as gradient decomposition. In particular, it leverages small auxiliary editing networks to transform the gradients obtained by standard fine-tuning into edits of weights in a pre-trained model. As gradients are generally high-dimensional objects, a low-rank decomposition of the gradients is utilized to achieve the transformation. Particularly, MEND parameterizes the gradient mapping functions as MLPs with a single hidden layer, such that a significantly small number of parameters are required, compared with the edited models. In this manner, MEND enables fast model editing that can operate on considerably large pre-trained language models.
%The authors evaluate MEND on language models including T5~\cite{raffel2020exploring}, GPT~\cite{radford2018improving}, BERT~\cite{devlin2018bert}, and BART~\cite{lewis2020bart}. The results demonstrate that MEND effectively edits the behavior of models with more than ten billion parameters.
Moreover, \textbf{KGEditor}~\cite{cheng2023editing} proposes to combine the benefits of memory-based methods and hyper-networks to ensure flexibility and further reduce computation costs. Particularly, KGEditor introduces an additional layer with the same architecture of FFN layers for storing knowledge. Then it constructs a hyper-network based on a bi-directional LSTM~\cite{hochreiter1997long} that encodes embeddings of triples. In this manner, KGEditor becomes an efficient way to edit knowledge graph embeddings.

\subsubsection{\textbf{Summary}}
Global optimization methods typically apply specific fine-tuning restrictions to regularize parameter updates, namely constrained fine-tuning strategies. This is to prevent overfitting and ensure the model's performance on the unedited knowledge. One crucial advantage of such strategies is its generality regarding the relevant knowledge, i.e., in-scope inputs $\mathcal{X}_e$ of edit $e$. As the global optimization affects all parameters in a language model, the relevant knowledge in it will also be edited, thereby generalizing to such knowledge. On the other hand, the high computation costs of fine-tuning all parameters also motivate researchers to propose intermediate fine-tuning strategies that leverage hyper-networks. Furthermore, global optimization methods are mostly model-agnostic, which means they can be applied to other editing methods. Nevertheless, such possibilities are less explored in the context of KME. In terms of the drawbacks, global optimization methods are suboptimal in maintaining the locality of edited models, as the optimization can easily influence unedited knowledge. Hence, it is crucial to achieve a balance between generality and locality when optimizing language models with specific constraints or intermediate designs.

%After a brief analysis of direct fine-tuning, we then discuss two main categories of methods in this field: meta-learning and subspace fine-tuning. These methods are reflected and extended in many other methods, such as SERAC~\cite{SERAC}, which scales MEND using memory-based approaches. Besides, many works of this nature are model-agnostic, and their validation, especially in the context of model editing and NLP-related tasks, is not thoroughly adequate. Future possibilities may include conducting relevant experiments and refining language model structures. 

\subsection{Local Modification}\label{sec:local}

\subsubsection{\textbf{Overview}}
To tackle the challenge of fine-tuning methods with respect to locality, extensive research has been conducted on the \textit{local modification} strategy for KME tasks~\cite{SERAC,yao2023editing}. These techniques originate from the concept of identifying and modifying specific relevant weights in a pre-trained model to achieve desirable outputs. The primary objective is to first locate the weights $\phi_{k}$ that store the knowledge in a pre-trained model $f_{\phi}$ regarding the input $x$. Afterward, by adjusting these weights, it becomes possible to generate the correct output $y^{*}$ from the same input $x$ without re-training or fine-tuning the whole model. Recently, researchers have generalized the local modification strategy to LLMs, where the efficiency of information updates for pre-trained LLMs can be substantially improved. Generally, the goal of the local modification strategy of KME can be formulated as a constrained optimization problem with refined constraints as follows:
\begin{equation} \label{eq:local}
        \begin{aligned}
        & \min_{\phi^{*}_{k}} \mathbb{E}_{e \in \mathcal{E}} \mathbb{E}_{x, y^* \in \mathcal{X}_e, \mathcal{Y}^*_e} \mathcal{L} (f^*_{\widebar{\phi}_{k}, \phi_{k}^{*}}(x), y^*), \\
        &\st f^*_{\widebar{\phi}_{k}, \phi_{k}^{*}}(x)=f(x),\ \  \forall x\in \mathcal{X}\setminus \mathcal{X}_\mathcal{E},\\
        & \text{where} \ \  \phi_k = L(f_{\phi}, \mathcal{E}),\ \widebar{\phi}_k = \phi \setminus \phi_k, \ f^*_{\widebar{\phi}_k, \phi^{*}_k}=M(f_{\phi}, \mathcal{E}).\\
    \end{aligned}
\end{equation}
Here $\phi^*$ denotes the edited weights related to the new knowledge, and $\widebar{\phi}_k$ denotes the unedited weights.
Eq.~(\ref{eq:local}) breaks down the local modification strategy for KME into two steps: (1) The \textit{locating step}, denoted by function $L$, locates the relevant weights $\phi_k$ in pre-trained model $f_{\phi}$ that store the obsolete information regarding the query $x$. (2) The \textit{editing step}, denoted by function $M$, edits the located weights $\phi_k$ into new weights $\phi_k^{*}$ such that the correct answer $y^{*}$ given the query $x$ can be generated by the model with $\phi_k^{*}$. By only updating a small fraction of model weights, the editing step avoids negatively influencing other irrelevant information, (i.e., $x \in \mathcal{X} \setminus \mathcal{X}_\mathcal{E}$). 

In the following subsections, we first introduce the concept of \textbf{knowledge neuron} in LLMs, which are specific neurons that store factual knowledge and can be activated to generate the desirable answer based on a certain query $x$. Then we discuss two local modification strategies for KME: (1) {the groundtruth-based strategies, which identify and edit knowledge neurons based on the supervision signal provided by the groundtruth}; (2) the prompt-based strategies, which locate knowledge neurons based on the input prompts.

%and prompt-based strategies, which find and edit knowledge neurons based on the properties of the outputs and the input prompt in a top-down and bottom-up manner, respectively.

\vspace{0.05in}
\noindent\textsf{\textbf{Knowledge Neurons.}}
LLMs pre-trained on large corpora can be viewed as databases that store factual and common-sense knowledge in the pre-trained model weights~\cite{gupta2023editing}. To update such knowledge by locally modifying the weights in the pre-trained LLMs, it is imperative to identify which weights store such information, i.e., locating the {knowledge neurons}. This can be challenging due to the complex transformer architecture of LLMs~\cite{bakker2022fine}. 

As described in Section~\ref{sec:transformer}, the transformer structure of LLMs consists of two primary types of layers, i.e., (1) the self-attention layer and (2) the point-wise feed-forward (FFN) layer, which is implemented as a two-layer multi-layer perceptron (MLP). 
Particularly, given a prompt $x$, the self-attention layers of the LLMs use the query vector of the last token and the key vectors of the previous tokens to calculate a weighted sum of their value vectors. Therefore, given the input $x$, these layers provide information about which previous tokens we should consider when generating the answer. Here we provide a simplified example for illustration. To answer the question \textit{``Who is the current president of the USA?''}, the self-attention layer indicates that the model should attend to words \textit{``president''} and \textit{``USA''}, i.e., $\textbf{v}_{president}$, $\textbf{v}_{USA}$, to determine the answer. This provides us with a start-up embedding $\textbf{h}^{start}$ to generate the answer token, which is the weighted sum of the values of the two attended words, i.e., $w_{1}\textbf{v}_{president} + w_{2}\textbf{v}_{USA}$. However, the information regarding the current president of the USA is not provided. In contrast, recent works~\cite{geva2021transformer,geva2022transformer,meng2022locating,meng2023mass} claim that the residual added to $\textbf{h}^{start}$ by the outputs of FNN layers, i.e., $\textbf{h}^{next} = \textbf{h}^{start} + \operatorname{FFN}(\textbf{h}^{start})$, injects the information \textit{``Biden''} to $\textbf{h}^{start}$ and leads to the generation of correct answers. Therefore, neurons in the FFN can be viewed as the \textbf{knowledge neurons} that store the factual knowledge. The role of FFN in storing knowledge can be theoretically analyzed by revisiting their formulation in Eq. (\ref{eq:transformer}), which we rewrite as follows:
\begin{equation}
\label{eq:geva}
\begin{aligned}
\text{SelfAtt}_i(\textbf{x})=\text{Softmax}\left(\textbf{q}_i \textbf{k}_i^\top\right) \textbf{v}_i, \quad
\text{FFN}(\textbf{h})=\text{GELU}\left(\textbf{h} \textbf{W}_1\right) \textbf{W}_2.
\end{aligned}
\end{equation}
Specifically, comparing the above two equations, we observe that the input $\textbf{h}$ to the FFN acts similarly to the query $\textbf{q}$ to the SelfAtt. Moreover, the weights of the first layer $\mathbf{W}_{1}$ can be viewed as the key $\mathbf{v}$, where $\operatorname{GELU}\left(\textbf{h} \textbf{W}_1\right)$ can be viewed as calculating an unnormalized attention score over the row vectors of $\textbf{W}_{2}$. Finally, the weights of the second layer $\textbf{W}_{2}$  can be viewed as the value (or the memory) that stores the knowledge, which can be retrieved according to the unnormalized weights calculated by the first layer. 

%After locating a subset of weights $\phi_{l}$ that store the information of obsolete information $y$ regarding $x$ (or can be updated to include new information $y^{*}$), we need to modify it into $\phi^{*}_{l}$ with the remaining part of the pre-trained weights unchanged (i.e., the $K$ function in Eq. (\ref{eq:local})), such that obsolete information $y$ can be removed from the LLM, while new information $y^{*}$ can be incorporated simultaneously. To edit individual weights based on identified, Dou et al. proposed to directly update the weights according to the located knowledge neuron that could be used to store the information, where the corresponding value in the matrix $W_{2}$

\subsubsection{\textbf{Groundtruth-based Strategies}} Based on the knowledge neuron view of the FFN layer weights in pre-trained LLMs, various {groundtruth-based} methods are proposed to locate and edit the pre-trained LLMs. Generally, they perform editing in a top-down manner, utilizing the supervision signal provided by the correct {groundtruth} $y^*$. As an exemplar work, \textbf{KD}~\cite{dai2022knowledge} proposes to change each weight $w^{(l)}_{i}$ (i.e., the $i$-th weight in the $l$-th layer of FFN) from 0 to the pre-trained value $\hat{w}^{(l)}_{i}$ and calculates the cumulative change in the probability of predicting the output $y^{*}$ with input $x$, where the weights with a high cumulative probability are considered relevant for knowledge regarding $y^{*}$. {DEPN \cite{wu2023depn} proposes a similar cumulative probability-based strategy to detect knowledge neurons that store privacy knowledge.} In contrast to locating and editing an individual weight ${w}^{(l)}_{i}$, \textbf{ROME}~\cite{meng2022locating} proposes to update an entire FFN layer to encode the new knowledge of $y^{*}$. Specifically, they view the second layer weights $\textbf{W}_{2}$ in the FFN layer in Eq. (\ref{eq:geva}) as a linear associative memory~\cite{kohonen1972correlation,anderson1972simple} in the form of $\textbf{K}\textbf{W}_{2} = \textbf{V}$, where the keys $\textbf{K}$ and values $\textbf{V}$ associated with $\textbf{W}_{2}$ can be directly calculated via pseudoinverse. With such a view of $\textbf{W}_{2}$ in the FFN layer, the optimization objective of updating it into $\hat{\textbf{W}}_{2}$ to encode new knowledge in the edit $e = (s,r,o\rightarrow o^{*})$ can be formulated as follows:
\begin{equation}
\label{eq:meng2022}
\min \|\textbf{K}\hat{\textbf{W}}_{2} - \textbf{V}\| \ \text{s.t.} \ \hat{\textbf{W}} \textbf{k}^*=\textbf{h}^*.
\end{equation}
Here $\textbf{k}^{*}$, which should encode the information of the subject $s$, is calculated by sampling multiple $x \sim \mathcal{X}_{e}$ and taking the average of the outputs from the first dense layer of the FFN. The target activation $\textbf{h}^{*}$ is calculated via optimizing the probability of outputting the correct answers $y^{*} \in \mathcal{Y}_{e}$ of the pre-trained LLM via the subsequent layers. Then, an efficient rank-one update is conducted on the weights $\textbf{W}_{2}$ according to Eq. (\ref{eq:meng2022}), such that after the update, the edited FFN layer can output the correct hidden representation $\textbf{h}^{*}$ conducive to the generation of the right answer $y^{*}$ from $\textbf{k}^{*}$. {The \textbf{ROME} framework has been shown to generalize to the large Mamba model \cite{sharma2024locating}}. Recently, \textbf{MEMIT} \cite{meng2023mass} proposes to further generalize the above editing strategy of the FFN layers of pre-trained LLMs to the mass editing of different knowledge. Particularly, with $u$ new edits $\{e_{1}, e_{2},\dotsc, e_{u}\}$ that are required to be updated in the weights $\textbf{W}_{2}$, the mass knowledge editing problem can be formulated as the following optimization problem:
\begin{equation}
\label{eq:memit}
\min \left(\sum_{i=1}^n\left\|\textbf{k}_i \hat{\textbf{W}}_{2} -\textbf{v}_i\right\|^2+\sum_{i=n+1}^{n+u}\left\|\textbf{k}^{*}_i \hat{\textbf{W}}_
{2} -\textbf{v}^{*}_i\right\|^2\right),
\end{equation}
where $\textbf{k}_{i}$, $\textbf{v}_{i}$ are the original key, value pairs associated with the weights $\textbf{W}_{2}$ (i.e., row vectors in matrices $\textbf{K}$, $\textbf{V}$ in Eq. (\ref{eq:meng2022})), whereas $\textbf{k}_{i}^{*}$, $\textbf{v}^{*}_{i}$ are the updated key, value pairs calculated from the $i$-th edit $e_{i}$ as with Eq. (\ref{eq:meng2022}). In addition, since multiple edits are required, the update is shared among different MLP layers, which is conducted in a top-down manner to prevent the potential issue of editing layers that could affect the ones that have already been edited. 
The residual for each edit is spread evenly over the range of the critical FFN layer. The strategy of residual attribution has recently been improved by \textbf{PMET} \cite{li2023pmet}, which adopts a square root strategy to spread residuals to bottom FFN layers such that more precise information can be conveyed to critical layers. {Furthermore, \textbf{EMMET} \cite{gupta2024unified} generalized ROME and MEMIT by formulating the mass knowledge editing problem as a preservation (of irrelevant knowledge)-memorization (of new knowledge) constrained optimization problem, where they derive closed form weight update formulae when the edit is exact, i.e., $\textbf{k}^{*}_i \hat{\textbf{W}}_
{2} = \textbf{v}^{*}_i$ instead of minimizing the MSE in Eq. (\ref{eq:memit}).}

{ From the application's perspective, to remove toxic knowledge of LLM, \textbf{DINM} \cite{wang2024detoxifying} identifies layers that store toxic knowledge with the discrepancy of toxic/non-toxic sequence embeddings, and uses the non-toxic samples to locally modify the weights of identified layers.}

\subsubsection{\textbf{Prompt-based Strategies}}

Tailored to characteristics of LLMs that provide answer $y^{*}$ based on the prompt $x$, the operation of locating and editing knowledge neurons can also be conducted in a bottom-up manner, which aims to change the prompt to detect neurons to be edited. Specifically, by masking out the key information and observing the difference of activations in the intermediate layers of the LLM, the weights that store the information regarding the query $x$ can be located and updated to store the new information $y^{*}$. For example, \textbf{ROME}~\cite{meng2022locating} proposes a corruption-and-restore based strategy to identify relevant layers (or their hidden output variables $\textbf{h}$) that store the information based on the prompt $x$. It first randomly masks the hidden representations of the key vectors $\mathbf{k}$ (as described in Eq. (\ref{eq:transformer})) of the tokens in the prompts from a certain intermediate layer of the pre-trained LLM. Then it calculates the reduced probability of predicting $y$ (i.e., the obsolete outputs) as the causal mediation effects of $x$ on $y$ mediated by $\textbf{h}$. Consequently, the weights in layers with large mediated effects are viewed as knowledge neurons that store the information of $y$. 
{
$\text{\textbf{MEMIT}}_{\text{\textbf{CSK}}}$ 
 \cite{gupta2023editing} extends the above corruption-based strategy to editing common sense knowledge. The authors argue that, different from the factual knowledge that can be directly retrieved by the subject $s$,
 %(transformed into the prompt $x$ as the input to the LLM), 
 the object $o$ and relation $r$ also matter for commonsense knowledge. Therefore, three types of corruption and edit locations, i.e., subject, verb, and object, are thoroughly analyzed, where the performance of editing commonsense knowledge can be improved. Moreover, \textbf{BIRD}~\cite{ma2023untying} studies the novel problem of bidirectional KME, which requires the edited model to possess reversibility. For example, 
if the phrase ``The capital of France is'' is edited to a counterfactual “London” within a model, it should logically be able to retrieve the inverse fact. That is, when presented with “London is the capital of,” the model should respond with “France” rather than “England”. Based on the strategy of ROME, BIRD introduces a novel objective that involves the bidirectional relationships between subject
and object in an edit. In this manner, the updated model weights can preserve reversibility by learning such information.
}

\subsubsection{\textbf{Summary}}

In this part, we introduce the local modification strategy for pre-trained LLMs for efficient updates of new information without adding new weights or optimizing the whole network. We start by analyzing the pivotal role of the point-wise feedforward layers, i.e., the FFNs, to store the factual information in pre-trained LLMs, with the knowledge neurons associated with the FFN layer thoroughly analyzed. {We then discuss the groundtruth-based strategies, which achieve the modification in a top-down manner, generally based on least squares objectives computed from the output $y$. We further discuss the prompt-based strategies, which conduct modifications in a bottom-up manner based on the input prompt $x$}. Nevertheless, the scalability and retainability of local modification methods lack further improvements, as the performance might deteriorate with more edits performed~\cite{meng2023mass}.

\section{Datasets}\label{sec:dataset}
Recently, multiple datasets have been established to facilitate the evaluation of KME methods, and we summarize the commonly-used datasets in Table ~\ref{tab:dataset} to benefit future KME research. Specifically, these datasets can be divided into two groups: \textbf{generation datasets} (i.e., textual output) and \textbf{classification datasets} (i.e.,  categorical output). The datasets are obtained from a variety of sources, including knowledge graphs, Wikipedia pages, crowd-sourced responses, etc., which are adapted by researchers to fit into the KME setting.

\subsection{Generation Datasets}
For generation datasets, the target is in the form of textual content that is required to be generated by LLMs.
Serving as pivotal resources to evaluate KME methods, most generation datasets are based on relational knowledge and used for assessing the ability of editing techniques to inject new factual knowledge. This is because relational datasets preserve more definitive answers for each input and thus are more convenient and precise for evaluation~\cite{zhang2024comprehensive,yao2023editing}. Specifically, these datasets are generally curated from the corresponding relational datasets to encompass diverse relational contexts, ranging from question-answer pairs to intricate multi-hop queries. Therefore, the most prevalent output format is an object to be predicted.

In this subsection, we present the most representative generation datasets, shedding light on their unique attributes, the nature of their content, and the specific challenges they present for evaluating KME methods on factual knowledge as follows:
%First, we provide a review of the most representative datasets for evaluating language model editing performance on factual knowledge. 

\begin{table}[!t]\centering
\caption{
Statistics of prevalent KME datasets, including generation and classification datasets.
}\vspace{-0.1in}
\renewcommand{\arraystretch}{1.25}
\setlength{\tabcolsep}{4.pt}
  \setlength{\aboverulesep}{0pt}
\setlength{\belowrulesep}{0pt}
\scalebox{0.75}{
\begin{tabular}{l|llllll}
\midrule[1pt]
\textbf{Dataset} & \textbf{Type} & \textbf{\# Train} & \textbf{\# Test} & \textbf{Input}& \textbf{Output}& \textbf{Used in}  \\ 
\hline
\multirow{1.}{*}{zsRE} & \multirow{1.}{*}{Relational} & \multirow{1.}{*}{244,173} & \multirow{1.}{*}{244,173} & \multirow{1.}{*}{Factual Statement} & \multirow{1.}{*}{Object} & 
\begin{minipage}[t]{3.9cm}
\cite{cao2021editing,mitchell2022fast,meng2022locating,SERAC,Transformer-Patcher,hartvigsen2022aging,meng2023mass,lee2022plugandplay,ni2023forgetting,e1,wang2024wise,p1,wang2024memoryllm,gangadhar2024modeleditingstandardfinetuning,jiang2024learning,yu2024melo,gu2024model,gupta2024unified}
\end{minipage} \\
CounterFact& Relational& N/A&21,919  &Factual Question & Object& 
\begin{minipage}[t]{3.9cm}
\cite{meng2022locating,IKE,meng2023mass,ni2023forgetting,e1,p1,wang2024memoryllm,p4,gangadhar2024modeleditingstandardfinetuning,sharma2024locating,hu2024wilke,gupta2024unified,yoon2024bigger}
\end{minipage}\\
WikiGen& Generation& N/A& 68k  & Wiki Passage& Continuation& \cite{mitchell2022fast}\\
T-REx-100/-1000 &Relational& N/A & 100/1,000 & Factual Statement &Object&\cite{CALINET,KAFT}  \\ 
ParaRel& Relational & N/A & 253,448 &Factual Question & Object& \cite{dai2022knowledge}\\
NQ-SituatedQA &  QA & N/A &67.3k &User Query & Answer& \cite{lee2022plugandplay,dai2023neural}\\
MQuAKE-CF/-T& Relational & N/A& 9,218/1,825&Multi-hop Question & Object& \cite{MQuAKE,gu2023pokemqa,p3,wang2024deepedit,e2,jiang2024learning}   \\
Hallucination& Hallucination&N/A&1,392&(Fake) Biography&Biography&\cite{hartvigsen2022aging, wang2024wise,yu2024melo}\\
MMEdit-E-VQA&Multimodal& 6,346& 2,093&Image \& Question&Answer&\cite{cheng2023can}\\
MMEdit-E-IC&Multimodal& 2,849& 1,000&Image&Description&\cite{cheng2023can}\\
ECBD&Relational&N/A&1000&Reference to Entity&Completion&\cite{EKP}\\
{Conflict Edit} & {Relational} & {N/A} & {7,500} & {Factual Statement} & {Object} & {\cite{li2024unveiling}} \\ 
{Round Edit} & {Relational} & {N/A} & {5,000} & {Factual Statement} & {Object} & {\cite{li2024unveiling}} \\ 
{UKE} & {Relational} & {N/A} & {2,478} & {Factual Question} & {Object} & {\cite{wu2024updating}} \\ 
{RippleEdits} & {Relational} & {N/A} & {5,000} & {Factual Question} & {Object} & {\cite{cohen2024evaluating,jiang2024learning}} \\ 
{VLKEB} & {Multimodal} & {5,000} & {3,174} & {Image} & {Description} & {\cite{huang2024kebench}} \\ 
{MLaKE} & {Multilingual} & {N/A} & {9,432} & {Question} & {Answer} & {\cite{wei2024mlake}} \\ \hline
FEVER& Fact Checking & 104,966 & 10,444& Fact Description& Binary Label & \cite{cao2021editing,mitchell2022fast,Transformer-Patcher,p4}\\
ConvSent& Sentimental &  287,802 &15,989 & Topic Opinion & Sentiment &\cite{SERAC} \\
Bias in Bio& Biographical &5,000&5,000& Biographical Sentence & Occupation & \cite{hernandez2023measuring}\\
VitaminC-FC&  Fact Checking & 370,653&55,197& Fact Description& Binary Label & \cite{SERAC}\\
SCOTUS&Categorization&7,400&931&Court Documents&Dispute Topic&\cite{hartvigsen2022aging,yu2024melo}\\
\midrule[1pt]
\end{tabular}}
\vspace{-.15in}
\label{tab:dataset}
\end{table}
\begin{itemize}[leftmargin=0.35cm]

\item \textbf{zsRE}~\cite{levy2017zero}: zsRE is one of the most prevalent Question Answering (QA) datasets extended and adopted by~\cite{cao2021editing,mitchell2022fast} for KME evaluation. zsRE is suitable for evaluating KME due to its annotations of human-generated question paraphrases, which allow researchers to assess the model resilience to semantically equivalent inputs. In zsRE, each relation is associated with a set of crowd-sourced template questions, such as ``What is Albert Einstein’s alma mater?''. Each entry cites a Wikipedia sentence, serving as the factual basis or provenance. The dataset also contains negative examples that are generated by pairing a valid question with a random sentence.
\item \textbf{CounterFact}~\cite{meng2022locating}: CounterFact is established to distinguish superficial alterations in the word selections and significant, generalized modifications in its foundational factual knowledge. Proposed in ROME~\cite{meng2022locating}, each entry in CounterFact originates from a related record in ParaRel~\cite{elazar2021measuring}, containing a knowledge triple and meticulously crafted prompt templates. It is important to note that all subjects, relations, and objects in this tuple are recognized entities in Wikidata~\cite{vrandevcic2014wikidata}.

%The dataset is obtained by selecting all hypotheses generated through beam search, excluding the top-1, ensuring that the high-probability outcomes are consistent with the model's distribution. The semantically equivalent inputs are achieved by back-translation, serving as the equivalence neighborhood. 

\item \textbf{WikiGen}~\cite{mitchell2022fast}: Firstly proposed in MEND~\cite{mitchell2022fast}, WikiGen consists of approximately 68k question-answer pairs, with a similar size to zsRE. Here, each question corresponds to a sentence randomly sampled from Wikitext-103, and each answer is a 10-token sample obtained from a pre-trained distilGPT-2 model~\cite{ma2021ditilgpt2}. It is noteworthy that greedy 10-token prediction of the base model only aligns with edit targets for less than 1\% of samples.

\item \textbf{T-REx-100 \& T-REx-1000}~\cite{elsahar2018t}: First used in CALINET~\cite{CALINET}, the authors adopt the classic relational dataset T-REx~\cite{elsahar2018t} for evaluating model editors by extracting factual triplets of varying sizes (100 and 1,000). Particularly, for each triplet, the authors insert the head and tail entities into the template in LAMA~\cite{petroni2019language} based on the relation they share, which results in two datasets with 100 and 1,000 facts, respectively, for the purpose of false knowledge detection. It should be noted that each fact in these datasets is represented by several paraphrased sentences.

\item \textbf{ParaRel}~\cite{elazar2021measuring}: ParaRel is an expert-curated dataset that comprises diverse prompt templates for 38 relations, sourced from the T-REx dataset~\cite{elsahar2018t}. Firstly used in KN~\cite{dai2022knowledge}, the authors insert the head entity into each relational fact and set the tail entity as a blank for prediction. To ensure a rich variety in templates,  relations with less than four prompt templates are excluded, resulting in 34 relations in total. Each of these relations, on average, preserves 8.63 distinct prompt templates, leading to a total of 253,448 knowledge-revealing prompts for 27,738 relational facts.

\item \textbf{NQ-SituatedQA}~\cite{kwiatkowski2019natural}: NQ (Natural Questions) is a comprehensive question-answering dataset originating from user searches. In PPA~\cite{lee2022plugandplay}, the authors utilize NQ as the source knowledge while excluding any outdated information as identified by SituatedQA~\cite{zhang2021situatedqa} to create a novel dataset NQ-SituatedQA. SituatedQA is a dataset containing questions within a subset of NQ that are dependent on specific time and location. The authors then incorporate the time-dependent QA pairs from this subset, annotated using the 2021 Wikipedia~\cite{vrandevcic2014wikidata} dump.

\item \textbf{MQuAKE}~\cite{MQuAKE}: MQuAKE is constructed from Wikidata~\cite{vrandevcic2014wikidata} for evaluating the effectiveness of KME methods on multi-hop questions. In particular, it is designed to assess whether the edited models can correctly answer questions generated by chains of facts in plain text. MQuAKE consists of two datasets. (1) MQuAKE-CF is a diagnostic dataset, specifically crafted to evaluate KME methods in the context of counterfactual edits. (2) MQuAKE-T focuses on temporal-based knowledge updates and is aimed at assessing the effectiveness of KME techniques in updating outdated information with contemporary factual data.

\item\textbf{Hallucination}~\cite{hartvigsen2022aging}: Firstly processed in GRACE~\cite{hartvigsen2022aging}, Hallucination is created from the dataset released in SelfCheckGPT~\cite{manakul2023selfcheckgpt}, where the authors prompt GPT-3 to generate biographies based on concepts extracted from WikiBio. The sentences are annotated regarding the factual accuracy, and hallucinations in them are identified. Then in GRACE, the authors process this dataset by further extracting Wikipedia summaries from WikiBio and thereby acquire the correct entry of each sentence. In this manner, every edit consists of a potentially false biography generated by GPT-3 as the prompt, and a ground truth output, which is the correct next sentence extracted from Wikipedia.
There exist 1,392 potential edits for test.

\item\textbf{MMEdit}~\cite{cheng2023can}: This dataset is the first to explore the possibility of editing multimodal LLMs. Specifically, MMEdit consists of two prevalent multimodal tasks: Visual Question Answering (VQA)~\cite{antol2015vqa} and Image Captioning~\cite{herdade2019image}. VQA involves developing algorithms that can analyze an image's visual content, comprehend questions asked in natural language about the image, and accurately respond to those questions. Image Captioning aims to understand an image and then generate a detailed and coherent natural language description of that image. To create dataset MMEdit, the authors utilize BLIP-2 OPT~\cite{li2023blip} and extract edit data from the evaluation datasets VQAv2~\cite{goyal2017making} and COCO Caption~\cite{chen2015microsoft}, specifically focusing on their suboptimal entries.

\item\textbf{ECBD}~\cite{EKP}: Based on the original dataset ECBD (Entity Cloze By Date)~\cite{onoe2022entity}, the authors process this dataset for a novel task, namely Entity Knowledge Propagation (EKP). The task aimed at updating model parameters to incorporate knowledge about newly emerged entities that are not present in the pre-training data of the language models. For instance, BERT~\cite{devlin2018bert}, trained in 2018, does not recognize ``COVID-19'' as it is a more recent entity. The processed dataset aims to provide evaluation for such a task with the help of definition sentences as input to update knowledge about new entities. The entities are taken from the date between 2020/01 and 2021/09 to ensure that they are not in training data. Each edit consists of a new entity, a description sentence, a probe sentence, and a ground truth completion. 

\item\textbf{VLKEB}~\cite{huang2024kebench}: {VLKEB (Large Vision-Language Model Knowledge Editing Benchmark) aims to address the unique challenges of editing large vision-language models, which face additional difficulties due to different data modalities and complex model components with limited data for LVLM editing. 
%To overcome the limitations of MMEdit~\cite{cheng2023can} in lacking high-quality synthesized images for evaluation,
%The existing LVLM editing benchmark, with its three metrics (Reliability, Locality, and Generality), is insufficient due to the low quality of synthesized evaluation images and the inability to assess whether models apply edited knowledge in relevant content. To overcome these limitations, 
VLKEB collects data from the multi-modal knowledge graph MMKG~\cite{liu2019mmkg} and extends the Portability metric for evaluation. With MMKG, VLKEB binds image data with knowledge entities, which can be used to extract entity-related knowledge for editing data. }

\item\textbf{MLaKE}~\cite{wei2024mlake}: {MLaKE (Multilingual Language Knowledge Editing) is proposed to evaluate the capability of KME methods in multilingual contexts and multi-hop reasoning across five languages: English, Chinese, Japanese, French, and German. MLaKE aggregates fact chains from Wikipedia in multiple languages and utilizes LLMs to generate questions in both free-form and multiple-choice formats. Notably, existing methods show relatively high generalization for languages within the same language family compared to those from different families. These findings underscore the need for advancements in multilingual knowledge editing.
}

\item\textbf{UKE}~\cite{wu2024updating}: { UKE (Unstructured Knowledge Editing) is proposed to evaluate the capability of KME methods in updating knowledge based on unstructured texts. Updating LLMs with texts appears to be a more realistic application, which is also more complex and difficult. The authors leverage subjects and objects in Wikidata~\cite{vrandevcic2014wikidata} and retrieve the corresponding Wikipedia article summaries as unstructured texts. The authors also utilize LLMs to generate summaries for edits in two existing datasets, CounferFact~\cite{meng2022locating} and MQuAKE-CF~\cite{MQuAKE}, to obtain unstructured texts.
}

\item\textbf{RippleEdits}~\cite{cohen2024evaluating}: {This dataset proposes a novel evaluation
criterion, which assesses the performance of KME methods on additional edits brought by an existing edit. In particular, injecting new knowledge (e.g., ``Jack Depp is the son of Johnny Depp'')  introduces a “ripple effect,” which necessitates the model to update related knowledge as well (e.g., ``Jack Depp is the sibling of Lily-Rose Depp''). Based on this, the authors construct RippleEdits, consisting of 5,000 edits with various types of ripple effects.
}
\item\textbf{Conflict/Round Edit}~\cite{li2024unveiling}: {This dataset pioneers in investigating the potential side effects of KME methods for LLMs. The proposed dataset and evaluation metrics underline two primary concerns: (1) Knowledge Conflict: Modifying sets of logically conflicting facts can amplify the existing inconsistencies within LLMs. (2) Knowledge Distortion: Altering model parameters to update factual knowledge can permanently disrupt the inherent knowledge framework of LLMs. The dataset is constructed from WikiData~\cite{vrandevcic2014wikidata} with specific logical rules.
}

\end{itemize}

\subsection{Classification Datasets}
Classification datasets are also widely adopted to evaluate the effectiveness of KME. These datasets consist of prompt-target pairs, where the target is a discrete label instead of a textual sentence. In the context of KME, these labels help ascertain the alignment of model performance with desired edits.
%, whether they are sentiment-based, fact-checking, or entaiKMEnt-related. 
The advantages of classification datasets also involve their preciseness in evaluation without the need to define the specific output space. In this section, we summarize notable classification datasets that have been tailored and leveraged for assessing KME techniques as follows:

\begin{itemize}[leftmargin=0.35cm]
\item \textbf{FEVER}~\cite{thorne-etal-2018-fever}: FEVER is a fact-checking dataset originally processed in KILT~\cite{petroni-etal-2021-kilt} for verifying factual knowledge in the form of binary classification. It necessitates the retrieval of sentence-level evidence to determine whether a claim is supported or refuted, and is widely used for evaluating the performance of KME. Specifically, FEVER excludes claims labeled as lacking sufficient information, as they typically do not provide any evidence to evaluate the claim. 

\item \textbf{ConvSent}~\cite{SERAC}: Firstly processed in SERAC~\cite{SERAC}, ConvSent is used to evaluate the capability of an editor to modify a dialog agent's sentiment about a particular topic without influencing its responses to other topics. ConvSent is obtained from a list of 15,000 non-numeric entities from zsRE~\cite{levy2017zero,cao2021editing}, combined with 989 noun phrases from GPT-3~\cite{brown2020language} for 15,989 topics. Particularly, for each entity, there are ten positive and ten negative sentiment completions, which can be noisy, from the BlenderBot model with 3B parameters~\cite{roller2021recipes}. The refined sentiment labels are achieved by a sentiment classifier~\cite{heitmann2020more} pre-trained on RoBERTa~\cite{liu2019roberta}. %

\item\textbf{Bias in Bios}~\cite{de2019bias}: Bias in Bios is a dataset originally proposed for fairness-related machine learning, containing approximately 397k short professional biographies of online individuals, which are not relatively famous. Each biographical sentence is assigned an associated occupation label for the described person. To adopt this dataset for evaluating the performance of KME methods, the authors of REMEDI~\cite{hernandez2023measuring} extract a single sentence, modify it to display only the person's first name, and then query the language model with the prompt that follows the structure: ``{Person} has the occupation of...''. Then they evaluate the relative probabilities of the language model assigned to 28 potential occupations, where the language model is considered to be correct if the ground-truth occupation is ranked top-1.

\item\textbf{VitaminC-FC}~\cite{schuster2021get}: Firstly processed in SERAC~\cite{SERAC}, VitaminC-FC is constructed based on a fact-checking dataset, VitaminC~\cite{schuster2021get}. Particularly, VitaminC consists of more than 400,000 evidence-claim pairs, each of which is assigned a binary label to indicate whether the evidence entails the claim. The dataset was gathered from over 100,000 Wikipedia
revisions that modify an underlying fact, along with additional synthetic ones. In SERAC, the authors convert VitaminC into a KME dataset by using the evidence as the edit descriptor and using claims from the same Wiki pages accordingly as in-scope samples.

\item\textbf{SCOTUS}~\cite{hartvigsen2022aging}: Firstly proposed in GRACE~\cite{hartvigsen2022aging}, SCOTUS is processed with label shift based on the dataset with the same name from Fairlex~\cite{chalkidis2022fairlex}. This classification task is to categorize U.S. Supreme Court documents from various decades into one of 11 topics. The topics are clustered based on the specific matter of dispute, such as Criminal Procedure,
Civil Rights, and First Amendment.
Due to the evolution of categorization rules over time, the label distributions in this dataset also shift. Specifically, 7.4k cases from 1946-1982 are used for training, and 931 cases from the 1991-2009 period are for test.

\end{itemize}

\begin{table}[!t]\centering
\caption{
Examples of different downstream applications of KME: Question Answering (QA), Fact Checking (FC), and Natural Language Generation (NLG).
}\vspace{-0.1in}
\renewcommand{\arraystretch}{1.3}
  \setlength{\aboverulesep}{0pt}
\setlength{\belowrulesep}{0pt}
\setlength{\tabcolsep}{4.8pt}
\scalebox{0.75}{
\begin{tabular}{l|llll}
\midrule[1pt]
\textbf{Task}& \textbf{Edit Descriptor $e$}  & \textbf{In-scope Input} $x\sim\mathcal{X}_e$& \textbf{Original Output} $y\sim\mathcal{Y}_e$ & \textbf{Target Output} $y\sim\mathcal{Y}_e^*$\\\hline
QA& (Kazakhstan, Captital,&What is the capital of& Astana&Nur-Sultan\\
&Astana$\rightarrow$Nur-Sultan) &  Kazakhstan?& & \\\hline
FC& (Marathon, Record, & Kipchoge holds the men's&True&False \\
&Kipchoge$\rightarrow$Kiptum)& marathon world record.&  & \\ \hline
\multirow{3}{*}{NLG}& (Jordan Poole, Play In,& Provide a short introduction &Jordan Poole entered  & In 2023, Jordan Poole transitioned  \\
& Warriors$\rightarrow$Wizards)&to Jordan Poole, describing&the Warriors' rotation&from the Warriors to the Wizards,  \\
&& his current position.& recently.&remarking a significant change.\\
\midrule[1pt]
\end{tabular}}
\label{tab:application}
\end{table}

\section{Applications}\label{sec:task}

KME can benefit multiple downstream applications with the ability to precisely and efficiently inject knowledge into pre-trained LLMs. In the following, we introduce several key applications of KME techniques in realistic scenarios, where intuitive examples are provided in Table~\ref{tab:application}.

%In current research, a diverse range of downstream tasks is used, but their objectives are mostly simple and straightforward:
%1) To verify whether the model remembers modified knowledge.
%2) To evaluate the model's performance on unmodified knowledge. Language model editing techniques are at the forefront of innovation in the Natural Language Processing (NLP) domain, especially in the realm of 

\subsection{Question Answering}
{\textbf{Background}.} Question Answering (QA) is a core NLP task that aims to comprehend queries posed by users in natural language and provide answers based on the encoded knowledge in the pre-trained language model~\cite{shin2020autoprompt}. Traditional models for QA are generally fixed in their knowledge, capturing only the information available at the training time of~\cite{petroni2019language, jiang2020can}. However, in our dynamic world, new information is generated incessantly, which necessitates the constant update of QA models~\cite{talmor2018commonsenseqa}. Fortunately, KME methods enable the modification of QA models to cater to specific questions without disrupting responses to other unrelated inputs. Therefore, with KME strategies, the QA model can be efficiently updated on the run, where the currentness of the model can be guaranteed. Consequently, language model editing techniques have found broad applications across a myriad of QA contexts with potentially distinct requirements~\cite{lee2022plugandplay}.

\noindent{\textbf{Existing Works}.} The QA task encompasses various aspects, such as conversational QA, definition-based QA, and notably, relation-based QA~\cite{pandya2021question}. Relation-based QA is primarily adopted as an evaluation benchmark as it necessitates the retrieval of precise real-world facts in response to queries. This particular emphasis on specific information retrieval renders relation-based QA especially conducive to the benefits of KME techniques. For example, \textbf{PPA}~\cite{lee2022plugandplay} introduces an innovative task of \textbf{CuQA} (Continuously-updated QA), which intentionally emphasizes recurrent, substantial edits for language models to constantly update them with new information. An important aspect of the {CuQA} task is to ensure that the existing pre-trained knowledge remains unaltered with the integration of new knowledge. Therefore, this property is one important evaluation to assess model editing in CuQA tasks. 
In \textbf{MQuAKE}~\cite{MQuAKE}, the authors innovatively propose a multi-hop QA task that involves answering questions generated by chains of facts in plain text. Specifically, the task requires edited models to infer implicit relations that can be several hops away from the objects in the edit. For example, when a language model is modified regarding the president of the USA, an ideal model should also authentically alter answers to ``Who is the son of the president of the USA'', which is a two-hop relation. Such a task is significantly more challenging as it necessitates the model to alter its reasoning results in addition to the original edit. Nevertheless, the proposed method MeLLo in MQuAKE still exhibits outstanding performance on this difficult task, demonstrating the potential of KME in generalizing edited knowledge to multi-hop relations.

\subsection{Fact Checking}

\noindent{\textbf{Background}.} Fact-checking (FC) is a pivotal task in journalism, information verification, and combating misinformation that aims to scrutinize and affirm the authenticity of claims, statements, or information in news articles, social media, and other media content~\cite{schuster2021get,galitsky2023truth}. In a world overwhelmed with ever-emerging information, fact-checking facilitates the trustworthiness in the sharing of distributed information, promotes information transparency, and aids individuals in making well-informed decisions~\cite{thorne-etal-2018-fever}. However, it is crucial to constantly update fact-checking models.
For instance, during the COVID-19 pandemic, initial understandings and guidelines about the virus evolved as researchers gathered more data~\cite{shahi2021exploratory}. A fact-checking model that cannot adapt to these rapidly changing facts would quickly become outdated and potentially spread misinformation, thereby requiring the application of language model editing.
By integrating KME techniques into fact-checking models to consistently update them with the latest information and facts, it becomes possible to ensure the currentness, trustworthiness, and accuracy of the model despite the persistent evolution of information.

% \subsubsection{\textbf{Motivation}.} Language model editing emerges as a particularly suitable technique for updating fact-checking models with new facts, which allows the model to adapt to the most recent information and events. This is particularly beneficial when dealing with dynamic information landscapes where facts and narratives change over time. For instance, during the COVID-19 pandemic, initial understandings and guidelines about the virus evolved as researchers gathered more data~\cite{shahi2021exploratory}. A fact-checking model that cannot adapt to these rapidly changing facts would quickly become outdated and potentially spread misinformation, thereby requiring the application of language model editing. On the other hand, the computational efficiency of language model editing also enables quicker updates without having to re-train or fine-tune the entire model from scratch. As a result, language model editing has become a promising strategy for continuously updating fact-checking models.

%However, there are also potential challenges. For instance, if not implemented meticulously, language model editing might introduce biases or distortions to the model, leading to unintended outputs. It's also imperative to ensure that while introducing new knowledge, the model doesn't "forget" or overshadow previously learned important information.

\noindent{\textbf{Existing Works}.} Recently, several works have proposed to apply KME techniques in fact-checking models. In~\cite{zhu2020modifying}, the authors first explore the potential of modifying specific factual knowledge within the transformer backbone of the fact-checking model while ensuring that overall model performance remains intact on facts irrelevant to the editing purpose. Particularly, they identify the critical components within the transformer backbones conducive to effective knowledge modifications. In \textbf{SERAC}~\cite{SERAC}, the authors propose to use evidence gathered from Wikipedia as edit descriptors to update potentially outdated knowledge in the model. The proposed method exhibits significant performance improvements over baselines and can be generalized to other in-scope inputs collected from the same Wikipedia page.

\subsection{Natural Language Generation}
\noindent{\textbf{Background}.} KME techniques are also promising to ensure the relevancy of the Natural Language Generation (NLG) task, which aims to generate coherent and contextually relevant content based on provided instructions~\cite{reiter_dale_1997}. Considering the rapid evolution of the global information landscape, it is essential for NLG models to remain up-to-date and ensure the accuracy of generated text while avoiding potentially false statements that may mislead the users. %Editing an NLG model presents a unique challenge to  the coherence of the generated texts, which is different from question-answering or fact-checking tasks. Specifically, the edited model should maintain the ability to handle contradiction, where the incorporated new evidence invalidates information in the existing article. That being said, the model should be faithful to both the original content and the new evidence and seek a balance between them to ensure coherence and truthfulness. 

\noindent{\textbf{Existing Works}.} In practice, several works have been proposed to apply KME methods to promote model performance in natural language generation tasks. For instance, \textbf{FRUIT}~\cite{logan2022fruit} proposes to update outdated Wikipedia articles according to the collection of new information about the article’s subject. Based on the T5 model~\cite{raffel2020exploring}, the authors utilize a compressed output format to eliminate the necessity of generating the entire update from scratch and promote thoughtful content structuring, which effectively handles the challenge of incoherence. In \textbf{MEND}~\cite{mitchell2022fast}, the authors apply their proposed method in the Wikitext generation task, where the edited model is required to produce credible 10-token extensions based on a provided Wikitext prefix~\cite{ma2021ditilgpt2}. With modification on multi-layer token-wise activations and gradients, the edited model presents higher coherence on the NLG task, which demonstrates the effectiveness of KME in generating target texts with richer information than QA or FC. 

%Natural Language Generation (NLG) is defined as the task of generating text from the underlying representation of information\cite{reiter_dale_1997}. These tasks encompass various applications such as text generation, machine translation, summarization, and dialogue generation. 

%Logan IV et al. \cite{logan2022fruit} introduce a novel text generation task called \textbf{FRUIT}, which aims to update an article to reflect new information on its subject. The authors provide a pipeline for extracting weakly supervised training and evaluation data from pairs of Wikipedia snapshots, and collected data for the years 2019-2020 and 2020-2021, as well as human-annotated gold evaluation data. They also provide benchmark results for popular generation systems and introduce EDIT5, a T5-based approach tailored to editing that establishes the state-of-the-art. The analysis shows that developing models that can update articles faithfully requires new capabilities for neural generation models, and opens doors to many new applications.

\section{Discussion}\label{sec:discuss}

%Although a number of works have been proposed for language model editing, there still remain several crucial aspects that are rarely considered by exiting works. By further research on the challenges, the researchers would find more inspiration and prompt better model performance. In the following, we provide the critical challenges that still remain to be solved.

\subsection{Challenges}
Despite the continual progress of works on KME, several critical aspects have been inadequately addressed by existing studies. Delving deeper into these challenges could offer researchers fresh insights and pave the way for the further advancement of the field. Consequently, we hereby outline the pressing challenges that await solutions in KME.

%In real-world applications, practitioners may usually encounter scenarios that require a massive number of knowledge edits. However,  An ideal solution should effectively address both objectives; however, the degree of the contradiction between these two objectives may vary based on the specific used strategies. 

\noindent{\textbf{{Trade-off between Locality and Generality.}}}
In KME, it is crucial to balance two objectives, \textit{locality} and \textit{generality} (as defined in Sec. \ref{sec:metric}), such that a higher edit success rate can be achieved with minimal negative influence on knowledge irrelevant to the edits. When editing a language model, a potential trade-off might emerge between these two desirable properties. As demonstrated in~\cite{yao2023editing}, local modification methods, such as MEMIT~\cite{meng2023mass} and ROME~\cite{meng2022locating} generally preserve a higher level of locality, as they locate precise locations of target knowledge to conduct the edition, which does not largely affect the unrelated weights. In addition, T-Patcher~\cite{Transformer-Patcher} points out that increasing the size of memory increases locality while decreasing the generality. These observations underscore the intricate balance between locality and generality. However, it remains challenging to tackle the trade-off problem and achieve a balance between these two desirable properties of KME methods.

\noindent{\textbf{Theoretical Analysis.}}
While many current KME studies focus on developing effective methods to enhance the editing performance regarding various desirable properties, there exists a notable gap between the practical application and the comparatively less discovered theoretical analysis. Recently, in~\cite{tanno2022repairing}, the authors provide theoretical support for the justification of identifying harmful training examples and editing the model by erasing the information from a Bayesian view.
LEACE~\cite{belrose2023leace} introduces an analytical framework that offers a theoretical perspective for the task of erasing target concept information from every layer in language models. In general, the benefits of incorporating theoretical analysis are multi-faceted. First, theoretical analysis provides a deeper understanding of the mechanics underlying KME, allowing for more principled approaches to editing. Second, a strong theoretical basis sets a solid foundation for future research, encouraging more rigorous and systematic exploration in the field of KME.
 However, to the best of our knowledge, there still does not exist any comprehensive theoretical analysis regarding the KME problem that involves novel knowledge. 
We hope that future research will enrich the theoretical discourse that can deliver profound insights into the substantial foundations of KME methods.

%. In real-world applications, the capability of handling a large number of edits ensures that edited models can adapt to a wide array of scenarios
\noindent{\textbf{Editing at Scale.}}
Another crucial property that hinders the practical application of KME is scalability — the ability of editing strategy to effectively perform a large number of edits simultaneously~\cite{mitchell2022fast}. For example, conversational systems~\cite{IKE} are expected to be constantly updated to incorporate an enormous number of global events and the information originating from them. However, as the number of applied edits increases, the coherence of language models is severely jeopardized, as multiple edits might contradict a broader spectrum of pre-existing knowledge in the models~\cite{wang2023easyedit}. This can lead to decreased editing performance in both locality and generality metrics~\cite{SERAC}. Although external memorization methods can alleviate such problems with a larger size of memories of additional parameters, they are still vulnerable if thousands of edits are required~\cite{meng2022locating}. Moreover, simply adapting single-edit techniques for a multi-edit environment by merely applying them sequentially has been demonstrated to be proven suboptimal~\cite{meng2023mass}. Therefore, the unique and intricate challenge of coherence renders editing at scale a formidable task.

\noindent{{\textbf{Unstructured Editing.}} KME faces significant challenges due to its evaluation strategies that focus on knowledge triples, e.g., $t=(s,r,o)$, which are not reflective of how real-world knowledge updates occur~\cite{zhang2024comprehensive,huang2024kebench}. In reality, updates are often found in unstructured texts such as news articles and scientific papers. To address this gap, a recent benchmark~\cite{wu2024updating}, namely UKE (Unstructured Knowledge Editing), is proposed to evaluate editing performance using unstructured texts as knowledge updates. The experimental results demonstrate significant performance declines of state-of-the-art KME methods. Notably, such a decline persists even with knowledge triplets extracted from unstructured texts. As such, it is imperative to develop more robust and adaptable methods that use unstructured texts for editing.}

\subsection{Future Directions}
Despite the recent achievements in the development of KME strategies for effective and efficient updating of new knowledge into LLMs, KME research is still in its emerging stage. Several promising directions could be pursued to further advance this field. Accordingly, we identify five inspiring and important open problems worthy of exploration in the future as follows:

\vspace{0.05in}\noindent\textsf{\textbf{Optimization-Free Editing.}}
Recently, prompt engineering has become a prevalent solution for modifying the behaviors of pre-trained LLMs in a human-preferable manner without the requirement of parameter update~\cite{dong2022survey}. For example, in-context learning provides task descriptions and/or demonstrations in the form of plain text to promote the model performance ~\cite{brown2020language}, which makes it a potentially more efficient and practical strategy for language models. We note that IKE~\cite{IKE} proposes a novel framework that relies on demonstration contexts for KME without parameter updating, which explicitly formats the demonstrations that can guide the language model to copy, update, and retain the prediction of different prompts. However, such a strategy is difficult to scale and usually has unsatisfactory retention. Therefore, it remains a crucial while challenging task to develop optimization-free KME methods.

\vspace{0.05in}\noindent\textsf{\textbf{Auto-Discovery of Editing Targets.}}
Current KME methods mainly rely on human expertise to identify and incorporate desirable knowledge into pre-trained LLMs~\cite{yao2023editing,wu2024updating,zhang2024comprehensive}. This approach is inherently labor-intensive and can incur significant costs, especially considering the vast and rapidly expanding new information needed to be integrated into language models. A promising future direction lies in the automation of the edits, which aims to identify, evaluate, and prioritize new knowledge that needs to be integrated from raw resources such as websites and social media. Through this strategy, the application of KME can be streamlined, rendering it more practical and adaptable in real-world scenarios. A straightforward solution would be crawling new knowledge and transforming it into a knowledge base, querying LLMs for each knowledge triple, and editing the wrong answer. However, such a strategy still lacks efficiency. Therefore, it remains a crucial task to discover editing knowledge from various resources without human effort.

%This would represent a transformative shift, transitioning from manual, human-driven edits to a more dynamic and autonomous knowledge update paradigm.

%In existing works, the desirable knowledge is still only achieved by humans. Such a process can be laborious and expensive, thus reducing the practicality of applying language model editing techniques. 

\vspace{0.05in}\noindent\textsf{\textbf{Continual Editing.}}
Current KME methods primarily consider one-step offline editing~\cite{cao2021editing,logan2022fruit}; however, such an approach is not aligned with real-world applications where models might continually encounter novel knowledge to be injected. 
%Notably, recurrent issues can disrupt user experiences across different platforms. 
For example, an online question-answering (QA) model may continually encounter reports of incorrect answers from end users, where the editing needs to be conducted on the run~\cite{Transformer-Patcher}. 
Therefore, an optimal KME technique should be capable of instantaneously and continuously rectifying emergent issues. We note that continual editing of pre-trained LLMs presents a unique challenge: preventing the edited models from forgetting or contradicting previous edits. Despite the inherent complexities, the persistent demand for continual editing in practice underscores the importance of solving this challenge.

%Nevertheless, due to the prevalent need of continual editing in realw-rld scenarios, it is also critical for langaugem model eting techniques to develop strategies to solve this problem.

%This realization propels us to reconceptualize the Model Editing (ME) task, broadening its scope to address a series of errors in real-time. We introduce this enhanced paradigm as the Sequential Model Editing (SME) task. The essential characteristics of a proficient sequential model editor are elaborated upon in Section 3.

\vspace{0.05in}\noindent\textsf{\textbf{Robust Editing.}}
An important direction for the advancement of KME lies in enhancing its robustness. In an era where misinformation spreads rapidly, it is urgent that edited models not only retain their accuracy but also resist adversarial attacks and misinformation~\cite{ganguli2022red}. Here, we should note that the concept of robustness extends beyond just maintaining factual accuracy; it involves fortifying the model against potentially adversarial external perturbations~\cite{perez2022red}. 
For example, if KME is maliciously applied to inject harmful knowledge into language models, the edited models can be easily transformed into tools for misinformation~\cite{alpaca}. Therefore, to prevent such cases, it is crucial for KME techniques to develop capabilities that can identify and counteract such unwanted inputs, thereby enhancing their resilience against adversarial actions.
%should have the ability to detect harmful knowledge and promote the resilience to attacks.
%With more and more LLMs open-sourced, it is critical to prevent the possibility of them being edited into harmful models. 
In practice, as the trend leans towards open-sourcing LLMs, it becomes ever more crucial to safeguard against potential manipulations that can turn these models harmful.

%While current techniques have made strides in precision, the dynamic nature of real-world data demands a reinforced emphasis on building steadfastness into language model editing methodologies. Moving forward, the fusion of accuracy with robustness will be a cornerstone for the next generation of language model editing techniques.

\vspace{0.05in} \noindent\textsf{\textbf{Editable Fairness.}} With the wide application of large language models (LLMs) to support decisions, the emphasis on fairness has grown significantly~\cite{wang2023large}, which requires LLMs to fairly treat people with diverse background~\cite{abid2021persistent}. However, LLMs trained on large datasets inevitably incorporate certain biases during this pre-training phase ~\cite{dong2019unified}. Fortunately, the precision and efficiency of KME techniques offer a promising solution to mitigate such biases and promote fairness in pre-trained LLMs. For instance, in a model designed to classify biographical sentences with occupation~\cite{de2019bias}, KME can be used to inject nuanced knowledge about a particular profession, guiding the model towards a more equitable understanding of individuals associated with that profession~\cite{hernandez2023measuring}. However, this remains a complex challenge, as fairness often entails considering disparate groups of individuals rather than specific people. This broader focus makes knowledge injection via KME a non-trivial task. Despite these difficulties, the enhancement of fairness in language models is paramount, and KME techniques present a promising avenue to achieve this goal.

\section{Conclusions}\label{sec:conclusion}
In this survey, we present a comprehensive and in-depth review of knowledge-based model editing (KME) techniques for precise and efficient updating of new knowledge in pre-trained LLMs. We first formulate the KME problem as a constrained optimization objective that simultaneously ensures the accuracy and retention of editing, which is general to encompass different KME strategies. We then provide an overview of the evaluation metrics for KME, which sheds light on the desirable attributes of edited models. Subsequently, we propose a structured taxonomy framework to systematically categorize existing KME techniques. Within each category, we outline the central challenges, elaborate on the %innovative strategies used by 
representative methods, and discuss their strengths and weaknesses. Furthermore, we summarize the datasets widely utilized to assess KME techniques, highlighting that certain techniques demand specific dataset structures for training or evaluation. To inspire researchers to devise more practical implementations, we also spotlight the real-world applications of KME techniques. Finally, we identify several potential challenges for future research and provide insightful directions that are conducive to further advancement of the field.

\begin{acks}
This work is supported by the National Science
Foundation under grants (IIS-2006844, IIS-2144209, IIS-2223769,
CNS2154962, and BCS-2228534), the Commonwealth Cyber Initiative awards (VV-1Q23-007, HV-2Q23-003, and VV-1Q24-011), the JP Morgan
Chase Faculty Research Award, the Cisco Faculty Research Award,
the Jefferson Lab subcontract, and the UVA 4-VA collaborative research grant.
\end{acks}

%%
%% The next two lines define the bibliography style to be used, and
%% the bibliography file.
\bibliographystyle{ACM-Reference-Format}
\bibliography{acmart}

%%
%% If your work has an appendix, this is the place to put it.
\appendix

%\section{Research Methods}

%\subsection{Part One}

%\subsection{Part Two}

\end{document}

%% file: category.tex
\usetikzlibrary{trees}
\usetikzlibrary{matrix, patterns, positioning}

\definecolor{connect-line}{RGB}{0,0,0}
\definecolor{middle-color}{RGB}{255,255,255}
% \definecolor{leaf-color}{RGB}{166,208,153}
% \definecolor{leaf-color}{RGB}{173,216,230}
\definecolor{leaf-color}{RGB}{255,255,255}
% \definecolor{line-color}{RGB}{166,208,153}
\definecolor{line-color}{RGB}{25,25,112}

\definecolor{yellow-defined}{RGB}{254,194,0}

\definecolor{black}{RGB}{0,0,0}

% \definecolor{external}{RGB}{230,75,53}
% \definecolor{global}{RGB}{7,105,82}
% \definecolor{local}{RGB}{207,134,0}

\definecolor{external}{RGB}{21,155,255}
\definecolor{global}{RGB}{255,55,55}
\definecolor{local}{RGB}{69,177,79}

\tikzstyle{yellow-root}=[draw=yellow-defined,
    rounded corners, minimum height=1em,
    fill=yellow-defined!40,text opacity=1, align=center,
    fill opacity=.5,  text=black,align=center,font=\footnotesize,
    inner xsep=3pt,
    inner ysep=1pt,
]

\tikzstyle{external-leaf}=[draw=external,
    rounded corners,minimum height=1em,
    fill=leaf-color!40,text opacity=1, align=left,
    fill opacity=.5,  text=black,align=left,font=\footnotesize,
    inner xsep=3pt,
    inner ysep=1pt,
]
\tikzstyle{external-middle}=[draw=external,
    rounded corners,minimum height=1em,
    fill=middle-color!40,text opacity=1, align=center,
    fill opacity=.5,  text=black,align=center,font=\footnotesize,
    inner xsep=3pt,
    inner ysep=1pt,
]
    
\tikzstyle{global-leaf}=[draw=global,
    rounded corners,minimum height=1em,
    fill=leaf-color!40,text opacity=1, align=left,
    fill opacity=.5,  text=black,align=left,font=\footnotesize,
    inner xsep=3pt,
    inner ysep=1pt,
]
\tikzstyle{global-middle}=[draw=global,
    rounded corners,minimum height=1em,
    fill=middle-color!40,text opacity=1, align=center,
    fill opacity=.5,  text=black,align=center,font=\footnotesize,
    inner xsep=3pt,
    inner ysep=1pt,
]

\tikzstyle{local-leaf}=[draw=local,
    rounded corners,minimum height=1em,
    fill=leaf-color!40,text opacity=1, align=left,
    fill opacity=.5,  text=black,align=left,font=\footnotesize,
    inner xsep=3pt,
    inner ysep=1pt,
]
\tikzstyle{local-middle}=[draw=local,
    rounded corners,minimum height=1em,
    fill=middle-color!40,text opacity=1, align=center,
    fill opacity=.5,  text=black,align=center,font=\footnotesize,
    inner xsep=3pt,
    inner ysep=1pt,
]

\forestset{
  custom edge/.style={
edge path={\noexpand\path[\forestoption{edge}, rounded corners](!u.parent anchor) -- (.child anchor)\forestoption{edge label}}
  }
}
\forestset{
  straight edge/.style={
    edge path={
      \noexpand\path [\forestoption{edge}] (!u.parent anchor) -- (.child anchor)\forestoption{edge label};
    }
  }
}

\begin{figure*}[ht]
\centering
\begin{forest}
  for tree={
edge={-, draw=connect-line, line width=1pt},
 % edge path={    \noexpand\path[\forestoption{edge}, rounded corners]    (!u.parent anchor) -- +(-1pt,0pt)  -| ([xshift=0pt].child anchor)\forestoption{edge label};  },
    grow=east,
    reversed=true,
    anchor=base west,
    parent anchor=east,
    child anchor=west,
    base=middle,
    font=\footnotesize,
    rectangle,
    line width=1.2pt,
    draw=connect-line,
    rounded corners,align=left,
    minimum width=2em,
    s sep=4pt,
    inner xsep=3pt,
    inner ysep=0pt,
  },
  where level=1{text width=4.5em}{},
  where level=2{text width=5em,font=\footnotesize}{},
  where level=3{font=\footnotesize}{},
  where level=4{straight edge, font=\footnotesize}{},
  where level=5{font=\footnotesize}{},
  [\textbf{KME} , yellow-root ,edge=external, 
        [External Memorization, external-middle, edge=external, text width=7.5em, fill=external!30
            [Memory-based, external-middle, text width=8.2em, edge=external, fill=external!30
                [ MeLLo~\cite{MQuAKE} {,} MemPrompt~\cite{MemPrompt}{,} IKE~\cite{IKE}{,}\\ Language Patch~\cite{Patch}{,} SERAC~\cite{SERAC}{,} KAFT~\cite{KAFT} 
, external-leaf, text width=15.8em,  edge=external ]
            ]
            [Extension-based, external-middle, text width=8.2em, edge=external, fill=external!30
                [CALINET~\cite{CALINET}{,} T-Patcher~\cite{Transformer-Patcher}{,} GRACE~\cite{hartvigsen2022aging}{,} \\
                COMEBA-HK~\cite{e1}{,} SWEA~\cite{e2}
                , external-leaf, text width=15.8em, edge=external]
            ]
        ]
        [Global Optimization, global-middle, edge=global, text width=7.5em, fill=global!30
            [Constrained Fine-tuning, global-middle, text width=8.2em, edge=global, fill=global!30
                [RecAdam\cite{chen2020recall}{,} Editable Training~\cite{sinitsin2020editable}{,} \\ PPA~\cite{lee2022plugandplay}{,} {Modifying-Memory}\cite{zhu2020modifying}{,} \\
                F-Learning~\cite{ni2023forgetting}{,} 
                  MELO~\cite{yu2024melo}{,} RECT~\cite{gu2024model}
,global-leaf, text width=15.8em, edge=global,] %ACT~\cite{wang2023cross}{,} OCGNN~\cite{wang2021one}{,}
            ]
            [Intermediate Fine-tuning, global-middle, text width=8.2em, edge=global, fill=global!30
                [ KGEditor~\cite{cheng2023editing}{,} KE~\cite{cao2021editing}{,} SLAG~\cite{hase2023methods}{,} MEND\cite{mitchell2022fast} 
                , global-leaf, text width=15.8em, edge=global]
            ]
        ]
        [Local Modification, local-middle, edge=local, text width=7.5em, fill=local!30
            [Groundtruth-based, local-middle, text width=8.2em, edge=local,fill=local!30
                [KD\cite{dai2022knowledge}{,} 
  ROME \cite{meng2022locating}{,} DEPN \cite{wu2023depn}{,} PMET \cite{li2023pmet}{,} \\ 
   MEMIT \cite{meng2023mass}{,} EMMET \cite{gupta2024unified}{,} DINM \cite{wang2024detoxifying} 
                , local-leaf, text width=15.8em, edge=local ] % \cite{xu2022more} 
            ]
            %edge path={\noexpand\path[\forestoption{edge}, rounded corners](!u.parent anchor) -- (.child anchor)}
            [Prompt-based, local-middle, text width=8.2em, edge=local,fill=local!30
                [MEMIT$_{\mathrm{CSK}}$ \cite{gupta2023editing}{,} BIRD \cite{ma2023untying}
                , local-leaf, text width=15.8em, edge=local]
            ]
        ]
    ]
\end{forest}
\vspace{-3mm}
\caption{The categorization of KME techniques for LLMs and the corresponding works.}
\vspace{-3mm}
\label{fig:categorization}
\end{figure*}